%% file: main.tex
\documentclass[11pt,a4paper]{article}
\usepackage{times,latexsym}
\usepackage{url}
\usepackage[T1]{fontenc}

\newcommand{\gl}[2]{%
\leavevmode\vtop{\hbox{#1}%
\hbox{#2\lower1.4ex\rlap{ }}}}

\usepackage[acceptedWithA]{tacl2021v1}

\usepackage{xspace,mfirstuc,tabulary}

\newif\iftaclinstructions
\taclinstructionsfalse %
\iftaclinstructions
\renewcommand{\confidential}{}
\renewcommand{\anonsubtext}{(No author info supplied here, for consistency with
TACL-submission anonymization requirements)}
\newcommand{\instr}
\fi

\iftaclpubformat %

\else

\fi

\usepackage{custom}
\input{macros}

\usepackage{subcaption}

\title{Can Language Models Learn Typologically Implausible Languages?}

\author{
    Tianyang Xu$^{a,b}$ \quad Tatsuki Kuribayashi$^c$ \quad Yohei Oseki$^d$  \\
    \textbf{Ryan Cotterell}$^a$ \quad \textbf{Alex Warstad}t$^{a,e}$\\
    $^a$ETH Z\"urich \quad
    $^b$Toyota Technical Institute at Chicago \quad 
    $^c$MBZUAI \\
    $^d$The University of Tokyo \quad 
    $^e$University of California San Diego  \\
    \texttt{sallyxu@ttic.edu} \quad
    \texttt{tatsuki.kuribayashi@mbzuai.ac.ae} \\
    \texttt{oseki@g.ecc.u-tokyo.ac.jp} \quad 
    \texttt{rcotterell@inf.ethz.ch} \quad 
    \texttt{awarstadt@ucsd.edu}
}

\date{}

\begin{document}
\maketitle
\begin{abstract}
Grammatical features across human languages show intriguing correlations often attributed to learning biases in humans.
However, empirical evidence has been limited to experiments with highly simplified artificial languages, and whether these correlations arise from domain-general or language-specific biases remains a matter of debate.
Language models (LMs) provide an opportunity to study artificial language learning at a large scale and with a high degree of naturalism.
In this paper, we begin with an in-depth discussion of how LMs allow us to better determine the role of domain-general learning biases in language universals.
We then assess learnability differences for LMs resulting from typologically \textit{plausible} and \textit{implausible} languages closely following the word-order ``universals'' identified by linguistic typologists.
We conduct a symmetrical cross-lingual study training and testing LMs on an array of highly naturalistic but counterfactual versions of the English (head-initial) and Japanese (head-final) languages.
Compared to similar work, our datasets are more naturalistic and fall closer to the boundary of plausibility. 
Our experiments show that these LMs are often slower to learn these subtly implausible languages, while ultimately achieving similar performance on some metrics regardless of typological plausibility.
These findings lend credence to the conclusion that LMs do show some typologically-aligned learning preferences, and that the typological patterns may result from, at least to some degree, domain-general learning biases.

\vspace{0.1em}
\hspace{.5em}\includegraphics[width=1.1em,height=1.1em]{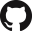}\hspace{.5em}\parbox{\dimexpr\linewidth-2\fboxsep-2\fboxrule}
  {\small \url{https://github.com/sally-xu-42/Typological_Universals}}
\end{abstract}

\section{Introduction}

\pgfdeclarelayer{background}
\pgfdeclarelayer{foreground}
\pgfsetlayers{background,main,depgroups,foreground}

\begin{table*}[t]
    \centering
    \small
    \begin{tabular}{ll}
        \toprule
        Correlation Pair & Example \\
        \midrule
        Original & \resizebox{1.5\columnwidth}{!}{%
    \begin{dependency}[theme = default, arc angle = 30]
       \begin{deptext}[column sep=0.3em, row sep=0em, font=\large]
          DET \& NOUN  \& AUX \& SCONJ \& DET \& NOUN \& ADP \& NOUN \& AUX \& VERB \& ADP \& PROPN \& ADP \& PROPN \\
          The \& fact \& is \& that \& the \& season \& of \& strawberries \& is \& running \& from \& July \& to \& August. \\
       \end{deptext}
       \deproot[edge height={1.8cm}]{3}{root}
       \depedge[edge style={font=\Large}, label style={font=\normalsize}, edge height={0.3cm}]{2}{1}{det}
      \depedge[edge style={font=\Large}, label style={font=\normalsize}, edge height={0.3cm}]{3}{2}{nsubj}
       \depedge[edge style={font=\Large}, label style={font=\normalsize}, edge height={1.5cm}]{3}{10}{cop*}
       \depedge[edge style={font=\Large}, label style={font=\normalsize}, edge height={1.2cm}]{10}{4}{mark}
       \depedge[edge style={font=\Large}, label style={font=\normalsize}, edge height={0.3cm}]{6}{5}{det}
       \depedge[edge style={font=\Large}, label style={font=\normalsize}, edge height={0.9cm}]{10}{6}{nsubj}
       \depedge[edge style={font=\Large}, label style={font=\normalsize}, edge height={0.6cm}]{6}{8}{nmod}
    \depedge[edge style={font=\Large}, label style={font=\normalsize}, edge height={0.3cm}]{8}{7}{case}
        \depedge[edge style={font=\Large}, label style={font=\normalsize}, edge height={0.3cm}]{10}{9}{aux}
        \depedge[edge style={font=\Large}, label style={font=\normalsize}, edge height={0.6cm}]{10}{12}{obl}
        \depedge[edge style={font=\Large}, label style={font=\normalsize}, edge height={0.9cm}]{10}{14}{obl}
        \depedge[edge style={font=\Large}, label style={font=\normalsize}, edge height={0.3cm}]{12}{11}{case}
    \depedge[edge style={font=\Large}, label style={font=\normalsize}, edge height={0.3cm}]{14}{13}{case}
       \end{dependency}
       } \\
        \cmidrule(lr){1-1} \cmidrule(lr){2-2}
        \corr{<V, O>} & \resizebox{1.5\columnwidth}{!}{%
    \begin{dependency}[theme = default, arc angle = 30]
       \begin{deptext}[column sep=0.3em, row sep=0em, font=\large]
          DET \& NOUN \& AUX \& SCONJ \& DET \& NOUN \& ADP \& NOUN \& ADP \& PROPN \& ADP \& PROPN \& AUX \& VERB \\
          The \& fact \& is \& that \& the  \& season \& of \&strawberries \& \textcolor{AccentBlue}{to} \& \textcolor{AccentBlue}{August} \& \textcolor{AccentBlue}{from} \& \textcolor{AccentBlue}{July} \& \textcolor{AccentRed}{is} \& \textcolor{AccentRed}{running}. \\
       \end{deptext}
       \depedge[edge style={font=\Large}, label style={font=\normalsize}, edge height={0.3cm}]{14}{12}{obl}
       \depedge[edge style={font=\Large}, label style={font=\normalsize}, edge height={0.6cm}]{14}{10}{obl}
       \wordgroup{2}{9}{10}{D2}
      \wordgroup{2}{11}{12}{D1}
        \wordgroup{2}{13}{14}{H}
    \end{dependency}
    } \\
        \corr{<Adp, NP>} & \resizebox{1.5\columnwidth}{!}{%
    \begin{dependency}[theme = default, arc angle = 30]
       \begin{deptext}[column sep=0.3em, row sep=0em, font=\large]
          DET \& NOUN \& AUX \& SCONJ \& DET \& NOUN \& NOUN \& ADP \& AUX \& VERB \& PROPN \& ADP \& PROPN \& ADP  \\
          The \& fact \& is \& that \& the \& season \& \textcolor{AccentRed}{strawberries} \& \textcolor{AccentBlue}{of}  \& is \& running \& \textcolor{AccentRed}{July} \& \textcolor{AccentBlue}{from} \& \textcolor{AccentRed}{August} \& \textcolor{AccentBlue}{to}. \\
       \end{deptext}
       \depedge[edge style={font=\Large}, label style={font=\normalsize}, edge height={0.3cm}]{13}{14}{case}
       \depedge[edge style={font=\Large}, label style={font=\normalsize}, edge height={0.3cm}]{11}{12}{case}
        \depedge[edge style={font=\Large}, label style={font=\normalsize}, edge height={0.3cm}]{7}{8}{case}
       \wordgroup{2}{14}{14}{D}
       \wordgroup{2}{13}{13}{D}
       \wordgroup{2}{12}{12}{D}
       \wordgroup{2}{11}{11}{D}
       \wordgroup{2}{8}{8}{D}
       \wordgroup{2}{7}{7}{D}
    \end{dependency} 
    }\\
        \corr{<Cop, Pred>} & \resizebox{1.5\columnwidth}{!}{%
    \begin{dependency}[theme = default, arc angle = 30]
       \begin{deptext}[column sep=0.3em, row sep=0em, font=\large]
          DET \& NOUN  \& SCONJ \& DET \& NOUN \& ADP \& NOUN \& AUX \& VERB \& ADP \& \& PROPN ADP \& PROPN \& AUX \\
          The \& fact \& \textcolor{AccentBlue}{that} \& \textcolor{AccentBlue}{the} \& \textcolor{AccentBlue}{season} \& \textcolor{AccentBlue}{of} \& \textcolor{AccentBlue}{strawberries}  \& \textcolor{AccentBlue}{is} \& \textcolor{AccentBlue}{running} \& \textcolor{AccentBlue}{from} \& \textcolor{AccentBlue}{July} \& \textcolor{AccentBlue}{to} \& \textcolor{AccentBlue}{August} \& \textcolor{AccentRed}{is}. \\
       \end{deptext}
       \depedge[edge style={font=\Large}, label style={font=\normalsize}, edge height={0.3cm}]{14}{9}{cop*}
       \wordgroup{2}{3}{13}{D}
       \wordgroup{2}{14}{14}{D}
    \end{dependency} 
    }\\
        \corr{<Aux, V>} & \resizebox{1.5\columnwidth}{!}{%
    \begin{dependency}[theme = default, arc angle = 30]
       \begin{deptext}[column sep=0.3em, row sep=0em, font=\large]
          DET \& NOUN  \& AUX \& SCONJ \& DET \& NOUN \& ADP \& NOUN \& VERB \& ADP \& PROPN \& ADP \& PROPN \& AUX \\
          The \& fact \& is \& that \& the \& season \& of \& strawberries \& \textcolor{AccentRed}{running} \& \textcolor{AccentRed}{from} \& \textcolor{AccentRed}{July} \& \textcolor{AccentRed}{to} \& \textcolor{AccentRed}{August} \& \textcolor{AccentBlue}{is}. \\
       \end{deptext}
       \depedge[edge style={font=\Large}, label style={font=\normalsize}, edge height={0.3cm}]{9}{14}{aux}
       \wordgroup{2}{9}{13}{D}
       \wordgroup{2}{14}{14}{D}
    \end{dependency} 
    }\\
        \corr{<Noun, Genitive>} & \resizebox{1.5\columnwidth}{!}{%
    \begin{dependency}[theme = default, arc angle = 30]
       \begin{deptext}[column sep=0.3em, row sep=0em, font=\large]
          DET \& NOUN  \& AUX \& SCONJ \& DET \& ADP \& NOUN  \& NOUN \& VERB \& ADP \& PROPN \& ADP \& PROPN \& AUX \\
          The \& fact \& is \& that \& the \& \textcolor{AccentBlue}{of} \& \textcolor{AccentBlue}{strawberries} \& \textcolor{AccentRed}{season} \& running \& from \& July \& to \& August \& is. \\
       \end{deptext}
       \depedge[edge style={font=\Large}, label style={font=\normalsize}, edge height={0.3cm}]{8}{7}{nmod}
       \wordgroup{2}{6}{7}{D}
       \wordgroup{2}{8}{8}{D}
    \end{dependency} 
    }\\
        \bottomrule
    \end{tabular}
    \caption{Illustrative examples of each of our counterfactual variants of English. Head phrases are colored red, and dependent phrases are colored blue. 
    In the \corr{<V, O>} example, we do not swap the copula and predicate due to readability, but these elements would be swapped in the actual dataset.
    The \corr{<V, O>} example demonstrates the reflective swapping (\corr{H D$_{1}$ D$_{2}$} $\rightarrow$ \corr{D$_{2}$ D$_{1}$ H}) explained in~\cref{ssec:general-swap}.
    }
    \label{tab:2}
\end{table*}

A fundamental goal in linguistics is to elucidate the universal properties underlying attested natural languages and to explain why some conceivable grammars but not others are widely attested.
Many typological universals and tendencies have been identified \citep{greenberg1963some,barwise1988generalized,Dryer1992TheGW,hyman2008universals}, but their causes are more elusive.
There is disagreement over whether typological patterns are caused by a learning bias that is language-specific \citep{chomsky1965aspects} or domain-general \citep{culbertson2016simplicity}, or even whether such a bias is the cause at all \citep{hahn2020universals}.
This debate has been difficult to resolve because we cannot manipulate variables during acquisition of a child's first language.
However, language models (LMs) have recently been advocated for as a convenient model for human learners that can enable large-scale controlled experiments on language acquisition \citep{warstadt2022artificial}.

Relatedly, a lively literature on counterfactual language learning in LMs has developed \citep[][i.a.]{ravfogel2019synthetic, hahn2020universals, White2021ExaminingTI, clark2023crosslinguistic, kallini-etal-2024-mission, kuribayashi-etal-2024-emergent}, sparking some debate.
\citet{chomsky2023nyt} criticized neural language models (LMs) as having little consequence for linguistic theory precisely because they can putatively learn both possible and impossible languages \citep{mitchell-bowers-2020-priorless}.
In response, \citet{kallini-etal-2024-mission} performed a set of experiments to test neural LMs' learnability of data with uncontroversially impossible properties as a natural language (e.g., lacking hierarchical structure), finding instead that LMs do indeed struggle with learning typologically impossible languages.

In this paper, we advance these debates by testing the learnability of typologically dispreferred languages that fall closer to the boundary of possibility.
The typological tendencies we study are those famously enumerated by \citet{greenberg1963some} and subsequently refined based on larger-scale typological studies \citep{Dryer1992TheGW}.
For example, languages with dominant subject-verb-object (SVO) order overwhelmingly have prepositions, while subject-object-verb (SOV) languages tend to have postpositions.
While previous work cited above has tested learnability of artificial languages with LMs, our approach to constructing counterfactual corpora has a unique combination of properties:
We aim to maximize naturalness by manipulating pre-existing natural language corpora and by iteratively annotating the counterfactual data and identifying and correcting corner cases.
We also target the decision boundary between typologically \textit{plausible} and \textit{implausible} languages by individually manipulating one specific grammatical property in each counterfactual corpus.
Finally, we balance biases due to the source language by symmetrically applying this procedure to a head-initial language (English) and a head-final language (Japanese).

In our experiments, we test the learnability of two types of LMs (autoregressive and masked) from scratch on each of our counterfactual languages. 
We evaluate learnability from multiple perspectives: (i) perplexity per token on the entire corpus, (ii) preferences on minimal pairs targeting the manipulated feature; and (iii) broad syntactic tests (BLiMP, \citeauthor{warstadt2020blimp}, \citeyear{warstadt2020blimp}; and JBLiMP, \citeauthor{someya-oseki-2023-jblimp}, \citeyear{someya-oseki-2023-jblimp}).
Our experimental results show that LMs often struggle to learn counterfactual, typologically implausible languages relative to minimally different natural languages. 
Thus we extend the findings of \citet{kallini-etal-2024-mission} on \textit{possible} vs. \textit{impossible} languages even closer to the boundary between \textit{plausible} vs. \textit{implausible} languages.
While we cannot entirely rule out confounds due to errors introduced in the creation of counterfactual corpora, these findings have important implications if they prove to be robust.
We also argue, contra \citet{chomsky2023nyt} and inspired by other recent arguments \citep{linzen2019can,warstadt2022what,wilcox2023using,constantinescuforthcominginvestigating} that learnability results from LMs can have important implications for our understanding of human language:
If Transformers, which lack language-specific learning biases, show a preference for typologically plausible languages, it is likely that humans have a similar learning preference as a result of domain-general learning biases.
Our results tentatively support this conclusion --- language-specific bias is not necessary, at least as a minimum requirement, to distinguish between typologically plausible and implausible word orders, pointing to a potential new line of evidence on a long-standing debate about the origins of linguistic typological patterns.

\section{Background}

\subsection{Typological Tendencies}
\label{ssec:tendencies}
What are all the conceivable grammars that human language could have?
While this might seem like an unanswerable question, linguistic theory gives us a particular kind of answer:
One of the key insights of modern linguistics is that natural language grammars can be viewed as instantiations of formal languages \citep{chomsky1956three}. 
Under this view, it becomes clear that there are conceivable classes of formal languages -- for example the regular languages -- to which no natural language belongs. 
But decades of research have shown that human languages occupy a much harder-to-define region within the high-dimensional space of possible grammars \citep{newmayer2005possible, chomsky1993theory}. 
Many generalizations have been made about the space of possible human languages, including generalizations about syntactic categories \citep{chomsky1965aspects}, quantifiers \citep{barwise1988generalized}, and phonology \citep{hyman2008universals}, to name just a few.
While some of these generalizations are true universals that no human language violates---for instance, no language has rules that require counting surface positions greater than two \citep{newmayer2005possible}---other generalizations are merely correlations of features that occur far more frequently than if features were sampled independently at random.
Thus, we can distinguish between \newterm{impossible} and \newterm{implausible} languages.

In the latter category, \citet{greenberg1963some} proposed a list of several dozen word order and morphological correlations based on a survey of 30 languages; 
for example, ``In languages with prepositions, the genitive almost always follows the governing noun, while in languages with postpositions it almost always precedes.''
Subsequently, \citet{Dryer1992TheGW} formulated a list of \newterm{correlation pairs}, that is, a list of pairs of morphosyntactic categories \corr{H} and \corr{D}\footnote{Syntacticians disagree on the correct generalization that characterizes these correlation pairs \citep{hawkins1983word,Dryer1992TheGW,kayne1994antisymmetry}, so there is no entirely theory-neutral description, perhaps besides ``verb-patterners'' and ``object-patterners''.
As suggested by our notation, the \corr{H} elements that pattern with the verb tend to be functional heads or lexical heads, while the \corr{D} elements that pattern with the object tend to be phrasal arguments or dependents. 
For example, the adposition is the functional head of an adpositional phrase.
While we tend to refer to these elements as heads and dependents, our study is predicated only on the existence of these correlation pairs, not the correct theoretical analysis.}  that tend to have the \textit{same} relative ordering as the dominant order of the verb and object, respectively, across a sample of 625 languages.
Following \citet{culbertson2015harmonic} we refer to languages (including most human languages) that follow these typological correlations as \newterm{harmonic} (i.e., plausible), while languages that violate them are \newterm{non-harmonic} (i.e., implausible).
A subset of these correlation pairs that we focus on in this paper is listed in Table \ref{tab:2}.%

\subsection{Learnability of Implausible Languages}
\label{sec:mechanisms}

Learnability has long been proposed as a primary mechanism behind typological universals and tendencies such as word order harmony.
This mechanism has an appealing story: Language evolves through re-analysis by child learners \citep{meillet1912levolution,cournane2017defence}, and re-analysis tends to favor easier-to-learn grammars, leading them to become more frequent on the scale of generations \citep{kirby2008cumulative}.
But why are some grammars harder or easier to learn in the first place?
Some scholars propose that humans have \newterm{language-specific biases}.
For instance, \cites{chomsky1965aspects} theory of \newterm{Universal Grammar} posited that humans have an innate language acquisition device that biases the learning of certain grammars.
This theory was later refined into the theory of \newterm{Principles and Parameters} \citep{chomsky1993theory}, which -- most relevant to the present discussion -- included a \newterm{Head Parameter} determining whether complements come to the left or right of their heads (p.~35).
Other scholars favor the view that \newterm{domain-general biases} are sufficient to explain some typological patterns. 
For instance, humans appear to have a \newterm{simplicity bias} across several domains of cognition \citep{chater2003simplicity,hsu2013language}, and such a bias could explain the preference for harmonic languages \citep{culbertson2016simplicity}, as harmonic grammars presumably have a shorter description length than non-harmonic ones.

From the empirical side, the evidence that human learning biases favor typologically plausible languages comes largely from artificial language learning experiments in laboratory settings.
Studies of this kind have shown that humans regularize novel grammatical rules in typologically plausible ways in the domains of phonology \citep{wilson2006learning} and morphology \citep{kam2005regularizing,fedzechkina2012language}.
Most relevant to the present discussion, a harmonic learning bias in artificial language learning has been found for English-speaking adults \citep{culbertson2012biases} and children \citep{culbertson2015harmonic}, as well as native speakers of cross-linguistically rare non-harmonic languages \citep{culbertson2020learning}.

\subsection{Counterfactual Language Paradigm}
Artificial or counterfactual language learning has also been widely applied to LMs in recent years \citep{ravfogel2019synthetic, hahn2020universals, White2021ExaminingTI, Hopkins-2022-towards, clark2023crosslinguistic, kallini-etal-2024-mission, kuribayashi-etal-2024-emergent}.
Whereas studies on human subjects are highly constrained by time, resources, and the limits of human attention, LMs can feasibly be trained extensively on artificial languages which can be highly complex, naturalistic, or formal.
Accordingly, the design space for these types of studies is large and comes with numerous trade-offs.
Specifically, we can distinguish the artificial language designs based on whether they take what we refer to as a \newterm{bottom-up} approach where a counterfactual corpus is generated from a manually specified lexicon and grammar; or a \newterm{top-down} approach where a naturalistic corpus is modified according to a set of rules.

At the extreme end of bottom-up approaches are studies that examine the learnability of different classes of formal languages for different neural network architectures, and therefore generate data potentially far outside the complexity class of natural language \citep{ebrahimi2020self,dusell2022stacks,hao-etal-2022-formal,deletang2023chomsky,borenstein2024easy,someya-etal-2024-targeted}.
\footnote{These empirical studies should be distinguished from theoretical studies that prove analytically which languages can be recognized by different architectures. See \citet{strobl2024formal} for an overview of that line of work.}
A slightly more natural approach is to design and generate texts from probabilistic context-free grammars inspired by those of natural language but which can violate specific typological properties.
These studies \citep{White2021ExaminingTI,kuribayashi-etal-2024-emergent} have yielded diverging results on whether the inductive biases of LMs align with those of humans.
However, bottom-up corpora massively simplify the problem of language learning and processing.
Naturalistic data contains a depth of constructions, statistical patterns, and errors that cannot practically be generated using a bottom-up approach.

The top-down approach achieves greater ecological validity by taking as a starting point a corpus that includes all the complexity of natural data, and performing controlled manipulations, often using constituency or dependency parses of the data.
One common approach uses parses to filter particular sentence types from a training corpus \citep{jumelet2018language,warstadt2022artificial,patil2024filtered,misra2024language}.
Other work applies rules to parses to modify sentences.
\citet{ravfogel2019synthetic} use gold parses from the Penn Treebank \citep{marcus1993penn} to create counterfactual versions of English with different agreement marking systems and each of the six possible dominant orders of subject, object, and verb.
\citet{hahn2020universals} create counterfactual dependency grammars by specifying for each arc label whether the dependent goes to the left or right and how close to the head it is placed relative to its sisters.
While this approach results in more ecologically valid counterfactual languages, it is also a noisy and difficult process to control.
Messy source data, annotation errors, or limitations of linguistic annotation systems mean that counterfactual corpora have more ungrammatical content (relative to the counterfactual grammar) than the original corpus.
Nonetheless, our study takes a top-down approach, while attempting to minimize and control for noise.

\section{Experimental Design}
\label{sec:hypotheses}
Our experiments test whether LMs show differences in learning natural languages with harmonic word orders compared to minimally different artificial languages with non-harmonic word orders (implausible languages).

\paragraph{The Independent Variable: Harmonic and non-harmonic languages}
We manipulate word order harmony using a top-down approach to counterfactual corpus generation. 
We modify naturally occurring corpora for languages with harmonic word orders by violating five specific Greenbergian correlation pairs, one at a time (see \cref{sec:counterfactual}).
For each
For each correlation pair, there are two types of harmonic languages (SVO with head-initial order, SOV with head-final order) represented by the natural corpora and two types of non-harmonic languages (SVO with head-final order, SOV with head-initial order) represented by counterfactual corpora.

\paragraph{The Dependent Variables: Measures for learnability}
There is no universally accepted definition or measure for learnability in the LM literature.
In this study, we investigated the learnability of counter-Greenbergian languages based on the learning trajectory of the LMs as well as their final performance after a certain period of training.
Given the concern that some counter-Greenbergian languages might eventually be \textit{learnable} for humans, one would naturally hypothesize that these tendencies could exist due to other learning barriers, such as \textit{learning efficiency}.
Therefore, we observed the learning trajectory of the counterfactual LMs across their checkpoints.
Details of our evaluation metrics and experimental results are shown in \cref{sec:experiment}.\looseness=-1

\paragraph{Addressing confounds: Symmetrical experimental design}
A key confound we try to avoid is that if we test on a fully head-initial language like English and make it head-final, the change in learnability can result from other factors than breaking the correlations, such as
(a) models' learning biases towards the head direction of a language, or 
(b) the amount of noise we induced during counterfactual corpus generation.
Our approach involves various ineliminable noise sources, including parser errors or ambiguities, punctuation removal prior to corpus editing, and the limitations of UD (universal dependencies) annotations.

We address (a) by conducting our experiments symmetrically with both a fully head-initial language and a fully head-final one.
We address (b) by reporting human validation scores, identifying parser ambiguities, and creating \vlm{Baseline} corpora variants that follow the same preprocessing steps of removing punctuation and lower-casing as applied to counterfactual corpora.

\section{Creating Counterfactual Languages}
\label{sec:counterfactual}
This section describes our procedure for creating counterfactual corpora by modifying natural sentences top-down. 
Implementation details and examples are further provided in \cref{app:details}.

\subsection{Swapping Greenbergian Correlation Pairs}
\label{ssec:general-swap}

\paragraph{Notation}
We denote a correlation pair using the notation \corr{<H, D>}, where \corr{H} is the \corr{Verb patterner} and is a mnemonic for \corr{head}, and \corr{D} is the \corr{Object patterner} and is a mnemonic for \corr{dependent}.
We use this notation to name a type of correlation pair by its syntactic categories (e.g., \corr{<Adp, NP>}) or to refer to a single instance of expressions belonging to the relevant categories (e.g., \corr{<in, the house>}).

\paragraph{Targeted Correlation Pairs}
Table \ref{tab:2} summarizes the selected subset of Greenbergian correlation pairs identified by \citet{Dryer1992TheGW} in our study.
As shown in \cref{tab:correlation_pairs}, we identify the five correlation pairs in dependency parses in the Universal Dependencies framework partly following \citet{hahn2020universals}.
While dependency arcs are a good start for identifying instances of \corr{H} or \corr{D}, they only connect two words, not entire phrases, and there is no one-to-one or even many-to-one correspondence between Universal Dependencies arcs \citep{demarneffe2021universal} and \cites{Dryer1992TheGW} correlation pairs.
For each language examined and each of the five correlation pairs, we implement a version of the swapping algorithm below to generate six distinct variants of a corpus with different word orders (Table~\ref{tab:2}).
\looseness=-1

\paragraph{Algorithm Overview}
The goal of our algorithm for creating counterfactual corpora is to swap the relative order of all instances of the relevant correlation pair within the input sentence. 
The word order is swapped at a span level.
That is, given a sentence $\bm w = [w_1, ..., w_n]$ and its dependency parse $p$, a word pair $(w_H, w_D)$ with a specific dependency type is first identified, and their spans $\bm s_H$ ($\bm s_D$) are determined as a continuous word sequence in $\bm w$ consisting of the identified word  $w_H$ ($w_D$) and its descendants\footnote{Strictly speaking, we use more different criteria to determine a word's span depending on the grammar of the language and the annotation, i.e. we do not always include all and only the descendants of the word.} in the dependency structure, i.e., $\bm s_h=[w_i,\cdots,w_H,\cdots,w_j]$; here, $1 \leq i\leq H\leq j \leq n$, and $w_H$ should be the head of the partial dependency structure of $s_H$. 
Then, the word order is swapped so that the relative position of $\bm s_h$ and $\bm s_d$ changes.
All the pairs of tokens (spans) that meet the criteria of an \corr{<H, D>}-pair in Table \ref{tab:correlation_pairs} are identified, and this span-swapping process is performed recursively (Algorithm \ref{alg:swap} in~\cref{app:details}).
Exceptions, additional conventions, and handling of coordination are covered in \cref{app:english}.

\paragraph{Handling Multiple Pairs}
In the case that multiple dependent spans share the same head $w_H$ in a sentence, we perform swapping by \emph{reflecting} the dependents around $w_H$.
In other words, we maintain the relative distance between \corr{H} and \corr{D}.
In an abstractive example of swapping ``\corr{H D$_{1}$ D$_{2}$}", the swapped order becomes ``\corr{D$_{2}$ D$_{1}$ H}".
In addition, since the dependency parse of a sentence exhibits a directed acyclic graph structure, and there might be nested correlation pairs, we perform a depth-first search over the sentence in our swapping algorithm (Algorithm \ref{alg:swap}).

\paragraph{Handling Japanese-Specific Issues}
Sometimes, a naive application of English implementation to Japanese does not work consistently due to differences in grammars and annotation conventions. 
For example, Japanese UD does not adhere to as rigid a notion of \emph{word} as English UD.
To make the swapping algorithm for both languages as similar yet correct as possible, we introduced some additional rules for the Japanese implementation (see \cref{app:japanese}).

\begin{figure}[t]
  \centering
  \includegraphics[width=1.0\linewidth]{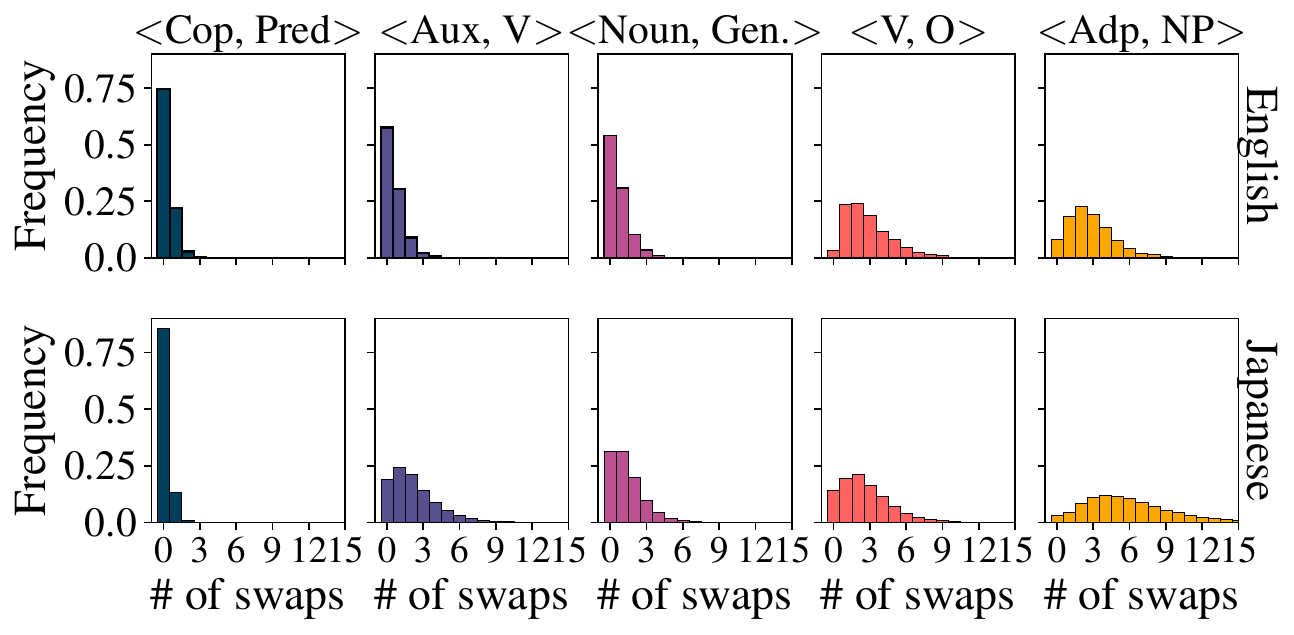}
  \caption{Histogram of the number of swaps per sentence for each counterfactual language.}
  \label{fig:freq}
\end{figure}

\paragraph{Statistics}
The frequency distributions of word-order swapping in a sentence for each correlation pair are shown in \cref{fig:freq}, which were estimated using a held-out set of LM training data (Wiki-40B).
The total number of swaps from lowest to highest is \corr{<Cop, Pred>}, \corr{<Aux, V>}, \corr{<Noun, Genitive>}, \corr{<V, O>}, and \corr{<Adp, NP>}.
Henceforth, the experimental results are reported in this order to facilitate interpretation of the results.

\begin{table}[t]
    \centering
    \scriptsize
    \tabcolsep=0.1cm
    \begin{tabular}{lrrrrrrrr}
    \toprule
    & \multicolumn{3}{c}{Train Data (En)} & BLiMP & \multicolumn{3}{c}{Train Data (Ja)} & JBLiMP \\
    \cmidrule(l){2-4} \cmidrule(l){5-5} \cmidrule(l){6-8} \cmidrule(l){9-9}
    Pair & Prec & Rec & Val & Val & Prec & Rec & Val & Val \\
     \cmidrule(l){1-1}  \cmidrule(l){2-2} \cmidrule(l){3-3} \cmidrule(l){4-4} \cmidrule(l){5-5} \cmidrule(l){6-6} \cmidrule(l){7-7} \cmidrule(l){8-8} \cmidrule(l){9-9}
    \corr{<Cop, P.>} & 59.1  & 54.2 & 4.4 & 4.9 & 55.0 & 55.0 & 4.8 & 4.9\\
    \corr{<Aux, V>} & 95.8 & 95.8 & 5.0 & 4.9 & 72.7 & 83.3 & 4.5 & 4.5 \\
    \corr{<N., Gen.>} & 80.0 & 80.0 & 4.8 & 5.0 & 81.0 & 81.0 & 4.8 & 4.9 \\
    \corr{<V, O>} & 74.4  & 73.4 & 4.3 & 4.6 & 85.9 & 81.6 & 4.2 & 4.3\\
    \corr{<Adp, NP>}  & 78.9 & 81.8 & 4.7 & 4.9 & 85.8 & 89.0 & 4.6 & 4.6 \\
    \bottomrule
    \end{tabular}
    \caption{Human validation results of counterfactual corpora. ``Prec,'' ``Rec,'' and ``Val'' denote precision, recall, and the averaged validation score indicated in the 5-point Likert scale.}
    \label{tab:validation}
\end{table}

\subsection{Human Data Validation}
\label{ssec:validation}
We conduct a human validation of our counterfactual corpora at several stages to ensure the validity of our swapping algorithm and iteratively improve our swapping algorithms.
Earlier iterations of validation were less formal, and resulted in changes to the swapping algorithm.
Below we describe the validation of our final counterfactual corpora.
While the swapping algorithm is not perfect, we believe that transparency about these flaws is an improvement over previous studies on top-down counterfactual corpora, none of which report any metric to evaluate the quality of their counterfactual corpora \citep{ravfogel2019synthetic,hahn2020universals,clark2023crosslinguistic}. 

\paragraph{Quantitative Evaluation}
Annotators manually list all $\corr{<H, D>}$ pairs that should be swapped for that sentence, according to their judgment. They compare this \textbf{gold} list to the \textbf{silver} list of all $\corr{<H, D>}$ pairs identified by the parser and swapped by the algorithm. We then compute the precision of the silver swaps ($\frac{\# \mathrm{correct\ silver}}{\# \mathrm{silver}}$) and the recall ($\frac{\# 
\mathrm{correct\ silver}}{\# \mathrm{gold}}$) over the entire annotated sentences. 

\paragraph{Qualitative Evaluation}
Annotators also subjectively assess the validity of each swapped sentence using a 5-point Likert scale (see \cref{app:annotation}).
This additional evaluation is motivated for several reasons: 
First, the quantitative evaluation unjustifiably favors mistakes that fail to identify a pair (which affects only recall) over mistakes where a silver pair is similar but not an exact match to a gold pair (which harms precision and recall).
Second, the silver string may sometimes be correct even if the identified pairs are not, i.e., some pairs are truly subjective due to ambiguity in the sentence or inevitable underspecificity in our annotation guidelines.
Third, errors can cascade, i.e., a single incorrect arc can lead to two (or more) errors arising from the words incorrectly connected and the words incorrectly \emph{not} connected.
Finally, some errors are intuitively less divergent from the counterfactual target (e.g., incorrectly resolving a prepositional phrase attachment) than others (e.g., misparsing a verb as a noun).

\paragraph{Annotators and Data}
One English native speaker and two Japanese native speakers annotated the gold word swap and the validity score for each sentence (each example was assessed by one annotator).
The annotators are all authors on the paper with PhD-level training in linguistics.
Our validation is mainly made on the training data for LMs (see~\cref{sec:experiment}), but we also conducted the qualitative evaluation part on sentences sampled from BLiMP/JBLiMP benchmarks, which are used in our LM evaluations~\cref{ssec:blimp}.
We sampled 120 sentences for \corr{<V, O>} and 40 sentences for the other correlation pairs for annotation, respectively, from the respective data sources, and thus 280 sentences are of validation target in each evaluation setting (e.g., English/Japanese LM training data).\footnote{We annotated an especially large number of sentences for the \corr{<V, O>} swap since it induced more diverse changes than the other correlation pairs.}
Notably, these validation targets include sentences without any target of respective swapping to properly estimate the precision of the algorithm.\footnote{When sampling LM training data to annotate, we balanced the data in each correlation pair to have 20 sentences with no silver swaps to better estimate the precision of the algorithm. Reported precision and recall reflect the distribution in the overall corpus, not the balanced sample.}

\paragraph{Results}
Table~\ref{tab:validation} shows the results.
The precision and recall of the word-sapping are typically above or near 80\%, and the average validity score on a 5-point scale is above 4.
Thus, we conclude that our word-swapping algorithm properly worked in most cases.
In addition, the 5-Likert scale scores are generally similar between LM training data and (J)BLiMP; thus, there are no issues specifically associated with the (J)BLiMP datasets, which include more complex or rare linguistic phenomena.
Though the swapping precision/recall for the \corr{<Cop, Pred>} part was particularly low, the validity scores are high.
This is due to frequent minor errors, typically in identifying the scope of the predicate in the copula construction.
For example, our algorithm converted a sentence ``\textit{he was active in the rsp student wing.}'' into ``\textit{he active \textcolor{red}{was} in the rsp student wing.},'' while human annotation was ``\textit{he active in the rsp student wing \textcolor{red}{was}.}''

\section{Model Training}
\label{sec:experiment}

\paragraph{Language Modeling}
To assess the inductive bias of both causal LMs and masked LMs, we duplicate our experiments with both GPT-2 small \citep{radford2019language} and LTG-BERT \citep{samuel2023trained} architectures.%
\footnote{LTG-BERT is a masked LM which resembles DeBERTa \citep{he2021deberta} with some additional optimizations. We choose this architecture as it is the basis for the model that won the BabyLM Challenge, a competition on data-efficient pretraining \citep{warstadt2023findings}.}
All models are trained for 12 epochs from scratch, and we examined three different random seeds for each setting.
\cref{app:lm} shows additional training details.

\begin{figure*}[t]
\centering
    \begin{subfigure}[c]{0.46\textwidth}
        \centering
        \includegraphics[width=\textwidth]{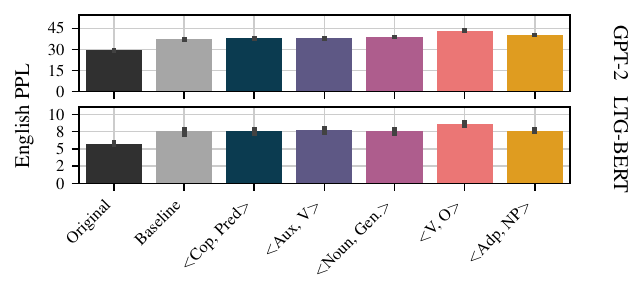}
    \end{subfigure}
    \begin{subfigure}[c]{0.46\textwidth}
        \centering
        \includegraphics[width=\textwidth]{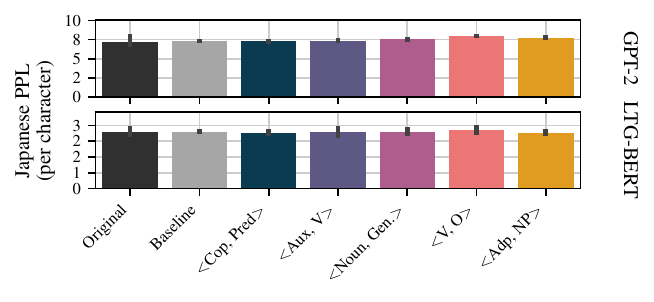}
    \end{subfigure}
    \caption{Final PPLs on respective heldout data for English-based counterfactual LMs (left) and Japanese-based counterfactual LMs (right). Error bars indicate standard deviation over three random seeds.}
    \label{fig:ppl-main}
\end{figure*}

\paragraph{Data}
We choose English and Japanese to perform our symmetrical (head initial/final$\rightarrow$final/initial) experiments.
Train, validation, and test splits consist of 100M words, 10M words, and 1M words, respectively. 
Token numbers are counted based on whitespace in English and MeCab~\cite{Kudo2005MeCabY} with the ipadic dictionary in Japanese, respectively.
These sentences are sampled from the English and Japanese parts of the Wiki-40B dataset \citep{guo-etal-2020-wiki}.
We choose Wikipedia data as the domain is similar to the data that the UD parsers were trained on, and thus we expect the resulting counterfactual corpora to be more accurate than would result from more developmentally plausible data such as child-directed speech.

We use Stanza to obtain dependency parses for every sentence in the corpora.
To avoid erroneous swapping, we removed (i) all punctuations from English and Japanese sentences; (ii) brackets (with their inside content) from Japanese sentences, i.e., typically rubi for Japanese Kanji; and (iii) sentences with lower-cased English words from the Japanese corpus.
We set two baseline models: (i) an \vlm{Original} model that is trained on our 100M Wiki-40B dataset without any preprocessing or swapping, and (ii) a \vlm{Baseline} model that is trained on the corpus with the preprocessing but without any swapping.
Comparisons between the \vlm{Original} and \vlm{Baseline} models function as a check for any unintended biases from our preprocessing.
Comparisons between \vlm{Baseline} and the other counterfactual LMs are of primary interest in how much counterfactual word order hurts language learning. 

\section{Results}

\subsection{Evaluation 1: Perplexity}
\label{ssec:trajectory}
\paragraph{Results}
We first compare the perplexities (PPLs) on the held-out data achieved by the LMs in each language, including counterfactual ones (Figure~\ref{fig:ppl-main}).\footnote{We report PPL per character for the Japanese results. This is necessary because the change in word order in different Japanese variants results in different token lengths due to the lack of whitespace word boundaries in Japanese.}
In the final epoch, the counterfactual LMs achieved similar PPL scores to the \vlm{Baseline} LMs.
However, if we look at the entire learning trajectory, learning appears to be slower for the counterfactual languages.
Note that the \vlm{Original} LMs also achieved PPL slightly better than \vlm{Baseline} but at an approximately similar scale; our preprocessing did not drastically change the language modeling task difficulty.
The \corr{<V, O>} variants tend to have slightly worse PPLs compared to \vlm{Baseline} and other counterfactual languages, which might be due to the fact that \corr{<V, O>} corpora have a large number of syntactically complex swaps (Figure~\ref{fig:freq}) and relatively worse swapping validity according to our human annotation (Table~\ref{tab:validation}).
Thus, the performance of LMs might plausibly reflect noise in the corpus as well as the difficulty of the (intended) grammar.

\paragraph{Statistical Tests}
We perform a paired Wilcoxon signed-rank test for each correlation pair in each source language by comparing six PPL scores of $\{\mathrm{GPT2, LTG\text{-}BERT}\}\times \{\mathrm{3\ seeds}\}$ from the corresponding counterfactual models and those from \vlm{Baseline} models.
Out of the ten settings of $\{\mathrm{En, Ja}\}\times \{\mathrm{5\ word\ orders}\}$, only one setting of English \corr{<V, O>} showed that the baseline model (real English) is significantly easier to learn ($p=0.03<0.05$) than the counterfactual one.
However, if we extend this analysis into learning trajectories, the statistical tests between 72 PPLs of $\{\mathrm{GPT2, LTG\text{-}BERT}\}\times \{\mathrm{3\ seeds}\}\times\{\mathrm{12\ epochs}\}$ can be made, and this yields that the real English is significantly easier to learn than the counterfactual one in all the five correlation pairs ($p<0.05$), while the real Japanese is \textit{not} significantly easier to learn than the counterfactual one in all the five correlation pairs ($p>0.05$).

\begin{figure*}[t]
    \begin{subfigure}[c]{0.5\textwidth}
        \centering
        \includegraphics[width=\textwidth]{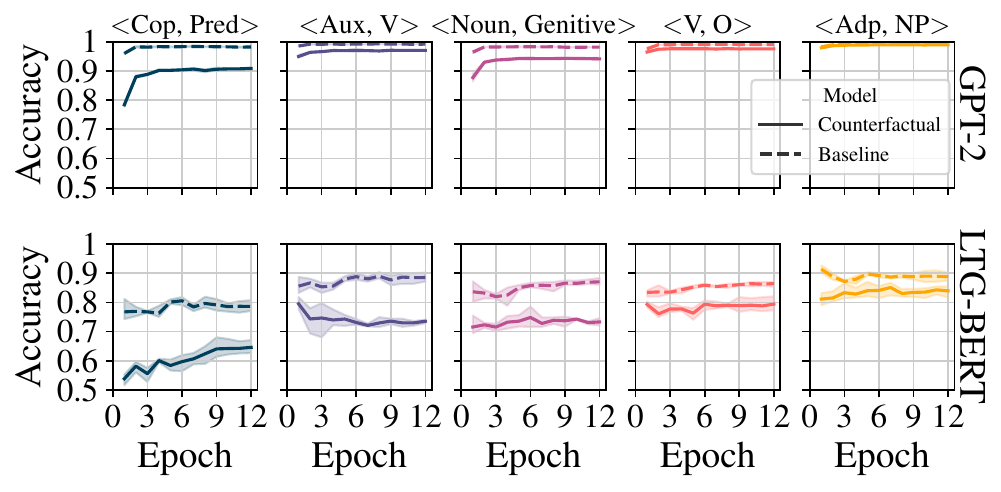}
    \end{subfigure}%
    \begin{subfigure}[c]{0.5\textwidth}
        \centering
        \includegraphics[width=\textwidth]{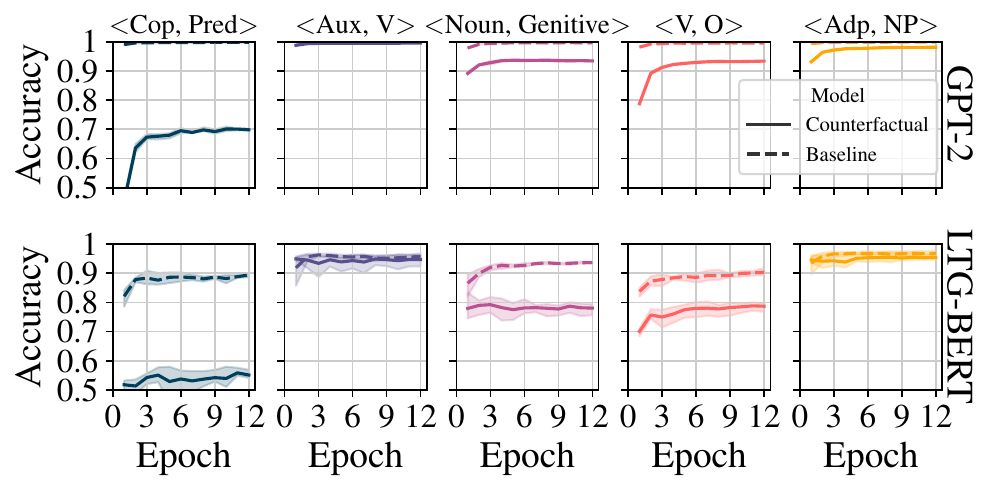}
    \end{subfigure}
    \caption{Performance trajectories for minimal pair comparisons targeting the counterfactual word order for counterfactual models and natural order for baseline model, for English-based LMs (left) and Japanese-based LMs (right). Shaded areas present standard deviation (SD) over three random seeds.}
    \label{fig:minimal}
\end{figure*}

\begin{figure*}[t]
    \begin{subfigure}[c]{0.5\textwidth}
        \centering
        \includegraphics[width=\textwidth]{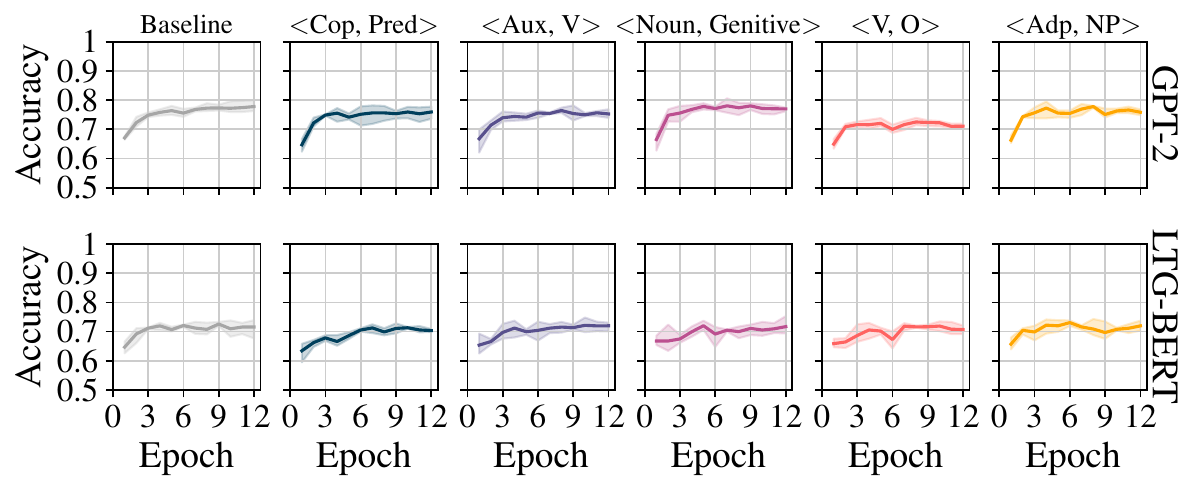}
    \end{subfigure}%
    \begin{subfigure}[c]{0.5\textwidth}
        \centering
        \includegraphics[width=\textwidth]{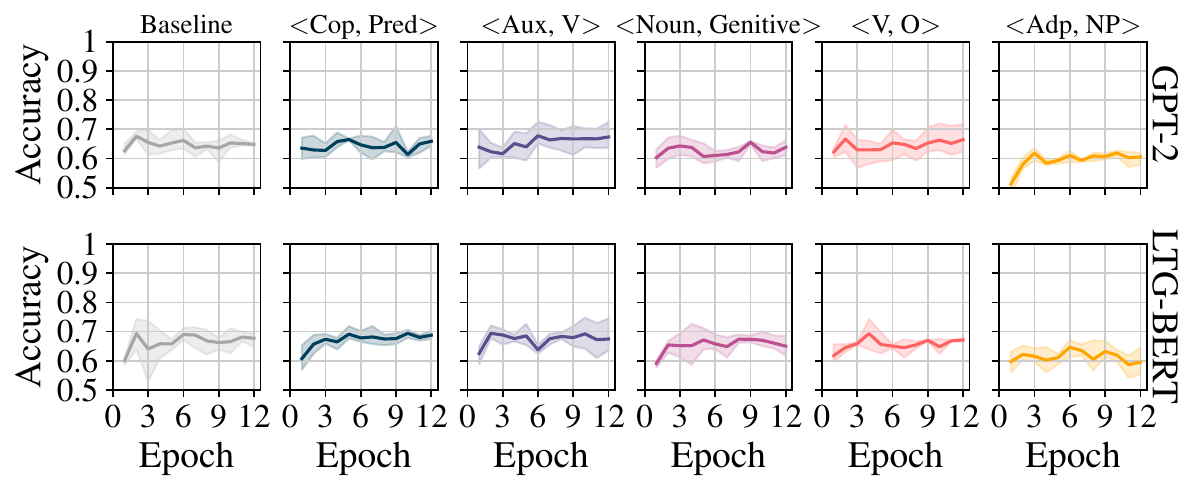}
    \end{subfigure}
    \caption{Performance trajectories of English-based counterfactual LMs in BLiMP (left) and Japanese-based counterfactual LMs in JBLiMP (right). Shaded areas present SDs over three random seeds.}
    \label{fig:blimp}
\end{figure*}

\subsection{Evaluation 2: Minimal Pair Preferences}
\label{ssec:minimal_pair}
\paragraph{Settings}
The previous evaluation measures PPL on all the tokens in the corpus; some of them are not necessarily related to our targeted word order change.
For a more targeted evaluation of the learnability of counterfactual Greenbergian word ordering, we design a binary task requiring selecting the right word order given two sentences containing at least one instance of a relevant correlation pair, differing only in whether the order of the elements in each pair is correct.
The task design is symmetrical between counterfactual and \vlm{Baseline} LMs; the correct option follows the counterfactual word order when evaluating counterfactual LMs, and vice-versa for the \vlm{Baseline} LMs.
To assess word order preference we compare the predicted probability (i.e., accumulated surprisal) of each sentence; that is, the option with a higher probability, i.e., lower surprisal, is regarded as preferred by LMs.
We report the accuracy in the binary task of selecting the correct word order.
The sentences were sampled from the held-out set of Wiki-40B data.

\paragraph{Results}
Figure~\ref{fig:minimal} shows the trajectory of accuracy during LM training.
All the counterfactual LMs prefer the correct word order over the incorrect one much more than random chance (accuracy of 0.5), which leads to our conclusion that LMs generalized well to counter-Greenbergian languages and learned the counterfactual ordering pattern successfully.
Nevertheless, in many settings, the \vlm{Baseline} LMs yielded higher accuracies than the counterfactual ones; thus, at least through the lens of this experiment, the real languages are usually easier to learn their word order for LMs.
However, this does not always appear to be the case as some counterfactual languages exhibit almost the same accuracies as the corresponding \vlm{Baseline} LMs, specifically for GPT-2.

\paragraph{Statistical Tests}
We perform a paired Wilcoxon signed-rank test for each correlation pair in each source language by comparing 72 accuracy scores of $\{\mathrm{GPT2, LTG\text{-}BERT}\}\times \{\mathrm{3\ seeds}\}\times\{\mathrm{12\ epochs}\}$ from the corresponding counterfactual models and those from \vlm{Baseline} models.
In all the ten settings of $\{\mathrm{En, Ja}\}\times \{\mathrm{5\ word\ order}\}$, the \vlm{Baseline} LMs exhibited significantly higher accuracies than the counterfactual LMs ($p<0.05$; in eight settings $p<$1e-12).\looseness=-1

\subsection{Evaluation 3: BLiMP \& JBLiMP}
\label{ssec:blimp}
\paragraph{Settings}
In addition to the minimal pair preference on Wiki-40B sentences (\cref{ssec:minimal_pair}), we further evaluate LMs on specific linguistic phenomena, ranging over morphology, syntax, and semantics, again using the minimal pair paradigm.
This evaluation tells us whether counterfactual word order has negative impacts on learning specific grammar rules not necessarily related to the swapped rule.
Specifically, we test LMs on a downsampled\footnote{We randomly sample 5 examples from each of the 67 BLiMP circuits, combine them into 12 BLiMP categories, and calculate the macro average accuracy over 12 categories.} version of BLiMP \cite{warstadt2020blimp} and JBLiMP~\cite{someya-oseki-2023-jblimp} benchmarks of minimal pairs for English and Japanese experiments, respectively.
For each counterfactual language, we also create a respective counterfactual version of BLiMP and JBLiMP by applying the same word-order swapping algorithm (\cref{sec:counterfactual}) to them.
Thus, each example in counterfactual (J)BLiMP consists of a pair of grammatically correct and incorrect sentences \emph{in the counterfactual language space}.
Notably, as demonstrated in~\cref{ssec:validation}, the accuracy of the word-order swapping algorithm was generally good even in BLiMP/JBLiMP datasets; this alleviates (but does not fully eliminate) the potential concern that these counterfactual versions of benchmarks are too noisy to estimate the model's linguistic knowledge.

\paragraph{Results}
We report the macro average of accuracy over the 12 BLiMP suites (or 9 JBLiMP suites).
Figure \ref{fig:blimp} shows the performance trajectory of LMs during training.
The \vlm{Baseline} trajectories are relatively similar or slightly better than those from counterfactual LMs, suggesting that counterfactual word order not drastically but slightly prevented LMs from acquiring grammatical knowledge.

\paragraph{Statistical Tests}
We performed a paired Wilcoxon signed-rank test for each correlation pair in each source language by comparing 864 accuracy scores of $\{\mathrm{GPT2, LTG\text{-}BERT}\}\times \{\mathrm{3\ seeds}\}\times\{\mathrm{12\ epochs}\}\times \{\mathrm{12\ BLiMP\ categories}\}$ from the corresponding counterfactual models and those from \vlm{Baseline} models.
In six of the ten settings of $\{\mathrm{En, Ja}\}\times \{\mathrm{5\ word\ orders}\}$, \vlm{Baseline} LMs exhibited significantly higher BLiMP accuracies than the counterfactual ones ($p<0.05$).\footnote{If we apply the Bonferroni correction, given that we performed statistical tests 30 times through our three experiments, the results could be more conservative, where baseline language yielded significantly higher BLiMP accuracies than four counterfactual languages ($p<0.0016=0.05/30$).}

\section{Discussion and Conclusions}

Our findings show that autoregressive and masked LMs have a consistent learning bias---with some notable exceptions---favoring harmonic languages over the nonharmonic counterfactual languages we examined.
Strikingly, the experimental results from Section \ref{ssec:minimal_pair} show that, for sentences involving the modified grammar rule, learning trajectories of the counterfactual languages lag behind those of the original language for every counterfactual language, model, and source language we examine.
The evaluations measuring PPL \cref{ssec:trajectory} and (J)BLiMP performance \cref{ssec:blimp} are more mixed, with counterfactual languages showing significantly worse performance across training only about half the time.

While these conclusions about LMs' learning biases are interesting in their own right, their implications for debates about linguistic typology are particularly important.
The role of modern LMs in linguistics and cognitive science has been a topic of much discussion and controversy \citep{pater2019generative,linzen2019can,baroni2022proper,warstadt2022what,lan2024large,wilcox2023using,piantadosi2023modern,katzir2023why,milliereforthcominglanguage,mcgrath2024how}.
Here, we will not rehearse all the details of this debate, but present a condensed account of how our experiments on LMs can inform ongoing debates about human language:

\citet{chomsky2023nyt} publicized aspects of this debate by claiming LMs learn impossible languages, and consequently have limited relevance to the study of human language.
\citet{kallini-etal-2024-mission} empirically test the first part of this claim in detail, showing that LMs display relative difficulty acquiring counterfactual versions of English with rules involving highly unnatural operations such as reversing strings and counting.
Our study furthers \citeauthor{kallini-etal-2024-mission}'s conclusions by showing that LMs continue to show a learning bias for typologically dispreferred counterfactual languages closer to the boundary between plausible and implausible.
However, we also take issue against the second part of \citeauthor{chomsky2023nyt}'s argument.\footnote{\citeauthor{kallini-etal-2024-mission} do claim that evidence from LMs is relevant to questions about the innate priors required for language learning (p.~14699) but do not fully spell out the argument that we give below.}

We contend that LMs, can help answer two questions about linguistic typology and acquisition, regardless (in some cases) of whether they show human-like biases.
First is the question of whether humans actually have a learning bias for harmonic languages. 
As discussed in \cref{sec:mechanisms}, the evidence in support of this conclusion from human subjects is limited somewhat due to the small scale and simplicity of the artificial languages employed.
Although LMs come with other limitations, the top-down approach to counterfactual language creation allows for naturalistic complexity and scale in the training data, providing a complementary line of evidence.
Our observation of a harmonic learning bias in LMs is powerful converging evidence adding to evidence from human studies that a learning bias for harmonic languages is real.
Given the LMs show this bias as well, there is less reason to doubt similar findings regarding humans.

Second is the question of whether a harmonic learning bias in humans is due to language-specific or domain-general priors.
The argument here is similar to that in several prior works \citep{clark2011linguistic,warstadt2022what,wilcox2023using,constantinescuforthcominginvestigating,kuribayashi-etal-2024-emergent}:
The Transformer architecture on which modern LMs are based \citep{Vaswani2017} is not specifically designed for language but is highly effective for domains as far reach as language, vision \citep{dosovitskiy2021an}, and protein sequences \citep{jumper2021highly}, suggesting that it relies on domain-general learning biases.
Thus, if one accepts that Transformes do show a harmonic bias, it follows that language-specific biases are not necessary to observe this phenomenon at least to some degree, and that should increase our credence in an explanation in terms of externally motivated domain-general biases in humans.
Furthermore, previous findings of harmonic bias in humans \citep[e.g., ][]{culbertson2012biases} might not be construed as evidence for language-specific bias in humans.

Importantly, evidence from this kind of experiment is relevant regardless of the result.
If we had found that Transformers did not show a harmonic bias, it would follow that such a bias is not a necessary consequence of the domain-general biases sufficient for language learning (at least assuming they learned the counterfactual languages successfully).
While it would still be possible in this counterfactual scenario that Transformers lack the relevant domain-general bias, we would nonetheless have increased our credence that humans might have some idiosyncratic learning mechanism which may well be language-specific.

It bears mentioning that even if humans do have a harmonic learning bias, other factors may still be equally if not more important to explain typological correlations.
Communicative pressures are another mechanism that might explain these phenomena, and extending our methods to test this mechanism is a promising avenue for future work.
\citet{hahn2020universals} and \citet{clark2023crosslinguistic} have both found that counterfactual languages perform worse than natural languages on measures of communicative efficiency, such as dependency length and uniformity of information density.
These measures can be straightforwardly applied to our counterfactual corpora, which employ both more targeted and syntactically informed manipulations than in those previous works. 

We must acknowledge an important limitation that tempers the force of our conclusions: Our manual validation shows that even our relatively careful approach to counterfactual language construction leads to numerous errors arising from parser errors.
Thus, it is possible that our findings may be due partially or entirely to increased noise in the counterfactual corpora, rather than inherent differences in learnability between the original and counterfactual grammars.
One defense against this unsatisfying conclusion is that on the PPL evaluation the final performance of counterfactual and original LMs are mostly not significantly different, suggesting that in the limit, the counterfactual languages are largely as predictable as the originals.
While it is true that the languages with the most noise according to our validity annotations, the \corr{<V,O>} languages, show the highest PPL, this pattern does not apply across other counterfactual languages.
We leave it to future work to explore alternative methods to reduce noise in naturalistic counterfactual corpora or to control for the amount of noise introduced by different forms of data manipulation.

Finally, while our study is a step forward in testing the learnability of counterfactual languages, it still leaves open many questions and avenues for future work.
Our conclusions are based only on two languages, so it will be important to try to replicate these results with more SVO and SOV languages, and also on languages with inconsistent VO ordering, such as German, though this direction will require input from many domain specialists and native-speaker linguists.
Future work should also study a wider variety of models as well as train models on more developmentally plausible data, such as dialogue data and child-directed speech.

To conclude, the rise of effective and efficiently trainable Transformer LMs has created the possibility of investigating the learnability of counterfactual languages at a scale and level of naturalism not possible with human subjects.
Through our emphasis on a syntactically sophisticated top-down approach to counterfactual language construction and the release of our code and models, we hope our work inspires further exploration of the diverse space of possible languages and deepens our understanding of the particular subspace that human languages occupy.

\section{Contribution Statement}
Tianyang Xu developed the primary codebase and conducted experiments. 
Tatsuki Kuribayashi contributed to the development of Japanese counterfactual corpora and Japanese validation processes. 
Alex Warstadt conceived of the study design, provided supervision throughout the implementation, and conducted the English validation.
Data visualization and statistical analysis were performed by TX, TK, and AW.
The first draft was written by TX, TK, and AW and edited by all authors.
Ryan Cotterell and Yohei Oseki offered feedback throughout the research process, and YO additionally contributed to Japanese validation.
\\ \noindent\textbf{Acknowledgements} AW acknowledges support from an ETH Postdoctoral Fellowship.
We are grateful to Ionut Constantinescu, Karen Livescu, Tiago Pimentel and Andreas Opedal for their insightful feedback and generous help.

\bibliography{custom}
\bibliographystyle{acl_natbib}

\newpage
\newpage
\appendix
\clearpage

\section{Implementation Details}
\label{app:details}
\subsection{English Policies}\label{app:english}
Here we provide the implementation details of the swapping algorithm for each correlation pair in the case of English experiments.
Unless otherwise specified in the next subsection, the same policy as English is adopted for Japanese.
Generally speaking, we identify correlation pair instances using the dependency arcs in \cref{tab:correlation_pairs}. However, there are numerous exceptions which we discuss below.

\begin{table}[ht]
\footnotesize
\centering
\begin{tabular}{lcl}
    \toprule
    \corr{H} & UD Relation & \corr{D} \\
    \cmidrule(l){1-1} \cmidrule(l){2-2} \cmidrule(l){3-3}
    \multirow{2}{*}{verb} & $\stackrel{\mathit{obj}}{\longrightarrow}$, $\stackrel{\mathit{iobj}}{\longrightarrow}$, $\stackrel{\mathit{obl}}{\longrightarrow}$ & \multirow{2}{*}{object} \\
    & $\stackrel{\mathit{cop*}}{\longrightarrow}$, $\stackrel{\mathit{ccomp}}{\longrightarrow}$, $\stackrel{\mathit{xcomp}}{\longrightarrow}$ & \\
    adposition & $\stackrel{\mathit{case}}{\longleftarrow}$ & NP \\
    copula verb & $\stackrel{\mathit{cop*}}{\longrightarrow}$ & predicate \\
    auxiliary & $\stackrel{\mathit{aux}}{\longleftarrow}$ & VP \\
    noun & $\stackrel{\mathit{nmod}}{\longrightarrow}$ & genitive  \\
    \bottomrule
\end{tabular}
\caption{Word orders of interest in Greenbergian correlation pairs and their associated Universal Dependencies, adopted mostly from \citet{hahn2020universals}. The asterisked \textit{cop*} is originally UD (universal dependencies) label \textit{cop} that we changed direction (lifted) during preprocessing, according to linguistic conventions.}
\label{tab:correlation_pairs}
\end{table}

\paragraph{Verbs and Objects}
We construe the \corr{<V, O>} correlation more broadly to refer to a verb on the one hand and its arguments and phrasal modifiers on the other.
In linguistic theory, there is no univerally agreed upon test for this notion of objectood.
To obtain a usable boundary for objects when swapping \corr{verb} and \corr{object}, we established five different selection criteria that identify objects with verbs based on their levels of connection, depicted in Figure \ref{fig:object}.
Each of the five criteria corresponds to a boundary, ranging from \emph{very tight} to \emph{very loose}, and we adopt the ``loose'' boundary for objects in our implementation.
Under this boundary, we treat all direct and indirect objects, prepositional objects, complement clauses and complement verb phrases, prepositional phrase adverbials, and non-finite adverbial clauses of a verb as objects in our implementation.

\tikzstyle{very tight}=[draw, fill=orange!90, text width=6em,
        text centered, minimum height=2.5em]
\tikzstyle{tight}=[draw, fill=orange!70, text width=6em,
        text centered, minimum height=2.5em]
\tikzstyle{medium}=[draw, fill=orange!40, text width=6em,
        text centered, minimum height=2.5em]
\tikzstyle{loose}=[draw, fill=orange!20, text width=6em,
        text centered, minimum height=2.5em]
\tikzstyle{very loose}=[draw, fill=orange!5, text width=6em,
        text centered, minimum height=2.5em]
\tikzstyle{brace}=[decorate,decoration={brace,raise=8pt,amplitude=0.2cm},black!70,very thick]
\def\horizondist{6}
\def\verticaldist{1.5}

\begin{figure}[ht]
\begin{center}
\scalebox{0.7}{%
\begin{tikzpicture}
    \node (dobj) [very tight] {direct object (NP)};
    \path (dobj)+(0,-\verticaldist) node (iobj) [tight] {indirect object (NP)};
    \path (iobj)+(0,-\verticaldist) node (iobjp) [medium] {indirect object (PP)};
    \path (iobjp)+(0,-\verticaldist) node (prepo) [medium] {prepositional object};
    \path (prepo)+(0,-\verticaldist) node (cc) [medium] {complement clause};
    \path (cc)+(0,-\verticaldist) node (cvp) [medium] {complement VP};

    \path (iobj)+(+\horizondist,+0.5*\verticaldist) node (ppadv) [loose] {PP adverbials (loc, temp)};
    \path (ppadv)+(0,-2*\verticaldist) node (nadvc) [loose] {nonfinite adverbial clauses};
    \path (nadvc)+(0,-2*\verticaldist) node (fadvc) [very loose] {finite adverbial clauses};

    \node (arguments) [above of=dobj] {\textbf{Arguments}};
    \path (arguments)+(+\horizondist,0) node (modifiers) {\textbf{Modifiers}};

    \coordinate (x) at (iobjp.north east);
    \coordinate (y) at (cc.south east);
    \coordinate (center) at ($(x)!0.5!(y) + (8pt+0.2cm,0)$);
    \coordinate (arrival) at ($(ppadv.west) - (0.2cm, 0)$);
    \coordinate (start) at ($(prepo.east) + (0.2cm, 0)$);
    \draw[->,black!70,very thick] (start) -- node[above,sloped,black,thick,text width=6em,text centered] {labeled as \textit{obl}}(arrival);

    \path (cvp.south west) ++(0.05*\horizondist,-1*\verticaldist) coordinate (colorbarstart);
    \draw [very tight, fill=orange!90] (colorbarstart) rectangle ++(1,0.3);
    \node [below, xshift=0.5cm] at (colorbarstart) {Very Tight};
    
    \path (colorbarstart) ++(1.8,0) coordinate (colorbarstart); %
    \draw [tight, fill=orange!70] (colorbarstart) rectangle ++(1,0.3);
    \node [below, xshift=0.5cm] at (colorbarstart) {Tight};
    
    \path (colorbarstart) ++(1.8,0) coordinate (colorbarstart); %
    \draw [medium, fill=orange!40] (colorbarstart) rectangle ++(1,0.3);
    \node [below, xshift=0.5cm] at (colorbarstart) {Medium};
    
    \path (colorbarstart) ++(1.8,0) coordinate (colorbarstart); %
    \draw [loose, fill=orange!20] (colorbarstart) rectangle ++(1,0.3);
    \node [below, xshift=0.5cm] at (colorbarstart) {Loose};
    
    \path (colorbarstart) ++(1.8,0) coordinate (colorbarstart); %
    \draw [very loose, fill=orange!5] (colorbarstart) rectangle ++(1,0.3);
    \node [below, xshift=0.5cm] at (colorbarstart) {Very Loose};

    \begin{pgfonlayer}{background}
        \path (dobj.west |- arguments.north)+(-0.5,0.3) node (a) {};
        \path (cvp.south -| cvp.east)+(+0.5,-0.3) node (b) {};
        \path[fill=blue!10, rounded corners, draw=black!50, dashed]
            (a) rectangle (b);
        \path (ppadv.west |- modifiers.north)+(-0.5,0.3) node (a) {};
        \path (fadvc.south -| fadvc.east)+(+0.5,-0.3) node (b) {};
        \path[fill=red!10, rounded corners, draw=black!50, dashed]
            (a) rectangle (b);        
    \end{pgfonlayer}
\end{tikzpicture}
}
\end{center}
\caption{Illustration of different levels of tightness when classifying verbal dependents as objects.}
\label{fig:object}
\end{figure}
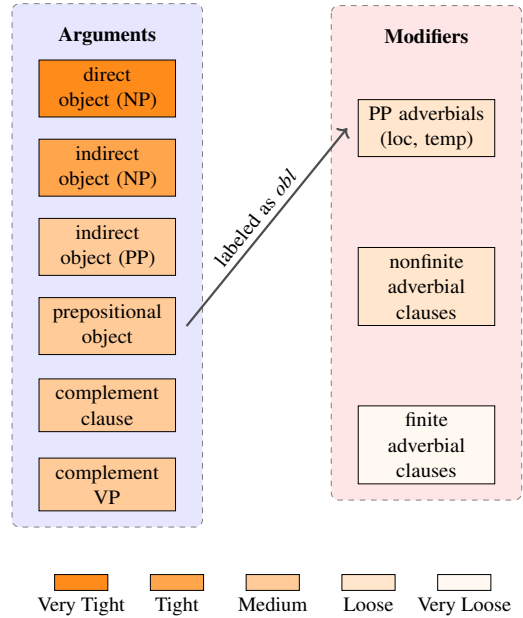

Mapping these linguistic constituents to UD relations, we use \deparc{obj, iobj, obl, cop, expl, xcomp} and \deparc{ccomp} as dependency arc labels to identify the \corr{<V, O>} pair.
Since the \deparc{ccomp} arc corresponds to both finite \& non-finite adverbial clauses, our approach depends on identifying an \deparc{nsubj} arc linked to the clause's main verb to differentiate between finite and non-finite adverbial clauses. 
We acknowledge that using the presence of a subject as the distinguishing factor might not be the best practice, given that the distinction between these clause types does not solely depend on having a subject, but it is an effective heuristic for most cases.

\paragraph{Adpositions and Noun Phrases}
POS tags \postag{NOUN, PROPN, NUM, PRON} for noun phrases and the UD arc label \deparc{case} identify the adposition and noun phrase word spans.
For compound adpositions, such as ``in front of'', we identify multiple \deparc{case} arcs one by one and swap accordingly.

\paragraph{Copula and Predicate}
The correlation pair \corr{<Cop, Pred>} is also included in \corr{<V, O>} pair in our formalization. 
In UD, the predicate is considered the head of the \deparc{cop} arc and all VP modifiers.
Following conventions in English syntax, we reverse the direction of the \deparc{cop} arc, making the copula the head of the predicate during preprocessing and transferring the VP modifiers to it before identifying both word spans using the \deparc{cop*}.

\paragraph{Auxiliary and Verb}
The \corr{<Aux, V>} pair is identified by UD relation \deparc{aux}.
We choose the associated verb phrase instead of a single verb for the word span of \corr{V} following conventions in English syntax.

\paragraph{Noun and Genitive}
The \corr{<Noun, Genitive>} pair is identified by UD relation \deparc{nmod}.
In English, however, possessive nominal modifiers are also labeled with \deparc{nmod}, such as \emph{John's book}, contrasting with \emph{book of John}. 
Thus we include an additional condition on the existence of ``of'' between a noun and its nominal dependents to identify genitives and exclude possessives.

To identify the span associated with the \corr{Noun}, we select all children preceding the \corr{Noun} and connected by \textit{nummod, compound, appos} and \textit{flat}, and all children between the \corr{Noun} and the genitive.
This choice is a heuristic developed through trial and error across several stages of annotation.

\subsection{Handling Coordination}

We also adopt a set of conventions regarding cases of coordination, illustrated in the table below using the correlation pair of \corr{<V,O>} as an example.

The first pair of rows illustrates cases where there is coordination of two dependents, which share a single head.
In such cases, we treat the pair of dependents plus the conjunction as a chunk that is swapped with the head.

The second pair of rows illustrates cases where there two head--dependent pairs are coordinated.
The dependency parse will have a \deparc{conj} arc between the two heads, and each head will have its own dependents. 
In such cases, we perform swapping for each head--dependent pair separately.

In the final two pairs of rows, we have two heads coordinated, with the second one having a dependent.
Importantly, in the first of these pairs, the dependent is shared by both heads, while in the second, the dependent belongs only to the second head.
Unfortunately, both sentences will receive the same dependency graph, so it is impossible to distinguish between these two cases.
We adopt the convention that the two heads are treated as a chunk when swapping with the dependent, although this inevitably leads to incorrect swaps in cases like last example below.
\vspace{1em}

{\centering\footnotesize
\begin{tabular}{ll}
\toprule
Constructions & Examples \\
\cmidrule(l){1-1} \cmidrule(l){2-2}
 $H \ D_1$ conj $D_2$ & we are students and teachers \\
$D_1$ conj $D_2 \ H$ & we students and teachers are \\
\cmidrule(l){1-1} \cmidrule(l){2-2}
$H_1 \ D_1$ conj $H_2 \ D_2$ & we like cats and love dogs \\
$D_1 \ H_1$ conj $D_2 \ H_2$ & we cats like and dogs love \\
\cmidrule(l){1-1} \cmidrule(l){2-2}
$H_1$ conj $H_2 \ D$ & we sing and dance in the park \\
$D \ H_1$ conj $H_2$ & we in the park sing and dance \\\cmidrule(l){1-1} \cmidrule(l){2-2}
$H_1$ conj $H_2 \ D$ & we dance and play tag \\
$D \ H_1$ conj $H_2$ & we tag dance and play \\
\bottomrule
\end{tabular}
}

\begin{table*}[t]
    \centering
    \small
    \begin{tabular}{ll}
        \toprule
        Correlation Pair & Example \\
        \midrule
        Original & \resizebox{1.5\columnwidth}{!}{%
    \begin{dependency}[theme = default, arc angle = 30]
       \begin{deptext}[column sep=0.3em, row sep=0em, font=\large]
          NOUN \& ADP  \& NOUN \& ADP \& NOUN \& ADP \& NOUN \& ADP \& VERB \& AUX \& NOUN \& ADP \& NOUN \& AUX \\
         Ichigo \& no \& kisetsu \& ga \& shichigatsu \& kara \& hachigatsu \& made \& tsudui \& teiru \& koto \& wa \& jijitsu \& dearu. \\
        Strawberry \& of \& season \& \texttt{NOM} \& July \& from \& August \& to \& runinng \& is \& that \& \texttt{TOP} \& fact \& is. \\
       \end{deptext}
       \deproot[edge height={3cm}]{14}{root}
       \depedge[edge style={font=\Large}, label style={font=\normalsize}, edge height={0.5cm}]{1}{2}{case}
      \depedge[edge style={font=\Large}, label style={font=\normalsize}, edge height={1cm}]{3}{1}{nmod}
       \depedge[edge style={font=\Large}, label style={font=\normalsize}, edge height={0.5cm}]{3}{4}{case}
       \depedge[edge style={font=\Large}, label style={font=\normalsize}, edge height={0.5cm}]{5}{6}{case}
       \depedge[edge style={font=\Large}, label style={font=\normalsize}, edge height={0.5cm}]{7}{8}{case}
       \depedge[edge style={font=\Large}, label style={font=\normalsize}, edge height={1.5cm}]{9}{5}{obl}
       \depedge[edge style={font=\Large}, label style={font=\normalsize}, edge height={1cm}]{9}{7}{obl}
    \depedge[edge style={font=\Large}, label style={font=\normalsize}, edge height={2cm}]{9}{3}{nsubj}
        \depedge[edge style={font=\Large}, label style={font=\normalsize}, edge height={0.5cm}]{9}{10}{aux}
        \depedge[edge style={font=\Large}, label style={font=\normalsize}, edge height={1cm}]{11}{9}{acl}
        \depedge[edge style={font=\Large}, label style={font=\normalsize}, edge height={0.5cm}]{11}{12}{case}
        \depedge[edge style={font=\Large}, label style={font=\normalsize}, edge height={1cm}]{14}{11}{nsubj}
    \depedge[edge style={font=\Large}, label style={font=\normalsize}, edge height={0.5cm}]{14}{13}{cop*}
       \end{dependency}
       } \\
        \cmidrule(lr){1-1} \cmidrule(lr){2-2}
        \corr{<V, O>} & \resizebox{1.5\columnwidth}{!}{%
    \begin{dependency}[theme = default, arc angle = 30]
       \begin{deptext}[column sep=0.3em, row sep=0em, font=\large]
          NOUN \& ADP  \& NOUN \& ADP \& VERB \& AUX \& NOUN \& ADP \& NOUN \& ADP \& NOUN \& ADP \& NOUN \& AUX \\
         Ichigo \& no \& kisetsu \& ga \& \textcolor{AccentRed}{tsudui} \& \textcolor{AccentRed}{teiru} \& \textcolor{AccentBlue}{hachigatsu} \& \textcolor{AccentBlue}{made} \& \textcolor{AccentBlue}{shichigatsu} \& \textcolor{AccentBlue}{kara} \& koto \& wa \& jijitsu \& dearu. \\
        Strawberry \& of \& season \& \texttt{NOM} \& running \& is \& August \& to \& July \& from \& that \& \texttt{TOP} \& fact \& is. \\
       \end{deptext}
       \depedge[edge style={font=\Large}, label style={font=\normalsize}, edge height={1cm}]{5}{9}{obl}
       \depedge[edge style={font=\Large}, label style={font=\normalsize}, edge height={0.5cm}]{5}{7}{obl}
       \wordgroup{2}{5}{6}{D}
       \wordgroup{2}{7}{8}{D}
       \wordgroup{2}{9}{10}{H}
       \end{dependency}
       } \\
        \cmidrule(lr){1-1} \cmidrule(lr){2-2}
        \corr{<Adp, NP>} & \resizebox{1.5\columnwidth}{!}{%
    \begin{dependency}[theme = default, arc angle = 30]
       \begin{deptext}[column sep=0.3em, row sep=0em, font=\large]
           ADP \& NOUN \& ADP \& NOUN \& ADP \& NOUN \& ADP \& NOUN \& VERB \& AUX \& ADP \& NOUN \& NOUN \& AUX \\
         \textcolor{AccentBlue}{No} \& \textcolor{AccentRed}{ichigo}  \& \textcolor{AccentBlue}{ga} \& \textcolor{AccentRed}{kisetsu}  \& \textcolor{AccentBlue}{kara} \& \textcolor{AccentRed}{shichigatsu} \& \textcolor{AccentBlue}{made} \& \textcolor{AccentRed}{hachigatsu} \& tsudui \& teiru \& \textcolor{AccentBlue}{wa} \& \textcolor{AccentRed}{koto} \& jijitsu \& dearu. \\
        Of \& strawberry  \& \texttt{NOM} \& season  \& from \& July \& to \& August \& runinng \& is \& \texttt{TOP} \& that \& fact \& is. \\
       \end{deptext}
       \depedge[edge style={font=\Large}, label style={font=\normalsize}, edge height={0.5cm}]{2}{1}{case}
       \depedge[edge style={font=\Large}, label style={font=\normalsize}, edge height={0.5cm}]{4}{3}{case}
       \depedge[edge style={font=\Large}, label style={font=\normalsize}, edge height={0.5cm}]{6}{5}{case}
       \depedge[edge style={font=\Large}, label style={font=\normalsize}, edge height={0.5cm}]{8}{7}{case}
        \depedge[edge style={font=\Large}, label style={font=\normalsize}, edge height={0.5cm}]{12}{11}{case}
        \wordgroup{2}{1}{1}{D}
        \wordgroup{2}{2}{2}{H}
        \wordgroup{2}{3}{3}{D}
        \wordgroup{2}{4}{4}{H}
        \wordgroup{2}{5}{5}{D}
        \wordgroup{2}{6}{6}{H}
        \wordgroup{2}{7}{7}{D}
        \wordgroup{2}{8}{8}{H}
        \wordgroup{2}{11}{11}{D}
        \wordgroup{2}{12}{12}{H}
       \end{dependency}
       } \\
               \cmidrule(lr){1-1} \cmidrule(lr){2-2}
        \corr{<Cop, Pred>} &  \resizebox{1.5\columnwidth}{!}{%
    \begin{dependency}[theme = default, arc angle = 30]
       \begin{deptext}[column sep=0.3em, row sep=0em, font=\large]
          NOUN \& ADP  \& NOUN \& ADP \& NOUN \& ADP \& NOUN \& ADP \& VERB \& AUX \& NOUN \& ADP \& AUX \& NOUN \\
         Ichigo \& no \& kisetsu \& ga \& shichigatsu \& kara \& hachigatsu \& made \& tsudui \& teiru \& koto \& wa \& \textcolor{AccentRed}{dearu} \& \textcolor{AccentBlue}{jijitsu.} \\
        Strawberry \& of \& season \& \texttt{NOM} \& July \& from \& August \& to \& runinng \& is \& that \& \texttt{TOP} \& is \& fact. \\
       \end{deptext}
    \depedge[edge style={font=\Large}, label style={font=\normalsize}, edge height={0.5cm}]{13}{14}{cop*}
        \wordgroup{2}{13}{13}{H}
        \wordgroup{2}{14}{14}{D}
       \end{dependency}
       } \\
               \cmidrule(lr){1-1} \cmidrule(lr){2-2}
        \corr{<Aux, V>} & \ \resizebox{1.5\columnwidth}{!}{%
    \begin{dependency}[theme = default, arc angle = 30]
       \begin{deptext}[column sep=0.3em, row sep=0em, font=\large]
          NOUN \& ADP  \& NOUN \& ADP \& NOUN \& ADP \& NOUN \& ADP \& AUX \& VERB \& NOUN \& ADP \& NOUN \& AUX \\
         Ichigo \& no \& kisetsu \& ga \& shichigatsu \& kara \& hachigatsu \& made \& \textcolor{AccentBlue}{teiru} \& \textcolor{AccentRed}{tsudui} \& koto \& wa \& jijitsu \& dearu. \\
        Strawberry \& of \& season \& \texttt{NOM} \& July \& from \& August \& to \& is \& running \& that \& \texttt{TOP} \& fact \& is. \\
       \end{deptext}
        \depedge[edge style={font=\Large}, label style={font=\normalsize}, edge height={0.5cm}]{10}{9}{aux}
        \wordgroup{2}{9}{9}{H}
        \wordgroup{2}{10}{10}{D}
       \end{dependency}
       } \\
               \cmidrule(lr){1-1} \cmidrule(lr){2-2}
        \corr{<Noun, Genitive>} &  \resizebox{1.5\columnwidth}{!}{%
    \begin{dependency}[theme = default, arc angle = 30]
       \begin{deptext}[column sep=0.3em, row sep=0em, font=\large]
          NOUN \& NOUN \& ADP \& ADP \& NOUN \& ADP \& NOUN \& ADP \& VERB \& AUX \& NOUN \& ADP \& NOUN \& AUX \\
         \textcolor{AccentRed}{Kisetsu} \& \textcolor{AccentBlue}{ichigo} \& \textcolor{AccentBlue}{no} \& ga \& shichigatsu \& kara \& hachigatsu \& made \& tsudui \& teiru \& koto \& wa \& jijitsu \& dearu. \\
        Season\& strawberry \& of \& \texttt{NOM} \& July \& from \& August \& to \& runinng \& is \& that \& \texttt{TOP} \& fact \& is. \\
       \end{deptext}
      \depedge[edge style={font=\Large}, label style={font=\normalsize}, edge height={0.5cm}]{1}{2}{nmod}
      \wordgroup{2}{1}{1}{H}
      \wordgroup{2}{2}{3}{D}
       \end{dependency}
       } \\
        \bottomrule
    \end{tabular}
    \caption{Counterfactual examples from our variants of the Japanese language. The word span of \corr{verb patterner} is colored red, and the word span of \corr{object patterner} is colored blue. 
    In the \corr{<V, O>} example, we omit the swapping regarding the cop dependency for the purpose of explanation and brevity.
    The \corr{<V, O>} example demonstrates the reflective swapping (\corr{H D$_{1}$ D$_{2}$} $\rightarrow$ \corr{D$_{2}$ D$_{1}$ H}) mentioned in~\cref{ssec:general-swap}.
    }
    \label{tab:japanese_examples}
\end{table*}

\subsection{Japanese-Specific Treatments}
\label{app:japanese}
Table~\ref{tab:japanese_examples} shows examples of counterfactual variants of a Japanese sentence.
The following paragraphs explain some treatments employed in modifying each word-order correlation pair in the Japanese language.

\paragraph{Verbs and Objects}
The Japanese language has a flexible word order, and the grammatical case of arguments is marked with a special marker rather than its word order~\cite{tsujimura2013introduction}.
However, these particles are sometimes omitted or overwritten by other particles, such as ``wa'' (topicalization marker; \texttt{TOP}) or ``mo'' (\textit{also}), making the grammatical relationships ambiguous superficially and leading to erroneous parser outputs.
To handle such errors, we employed several heuristic rules on top of the parser output to improve the accuracy and consistency of the word-order swapping algorithm:

\begin{itemize}
\item If a word has a \deparc{nsubj} dependency AND the nominative case marker ``ga,'' the word is treated as a subject (i.e., the word is not swapped).
\item If a word has a topicalization marker ``wa,'' the word is not swapped.
\item The other arguments with the \deparc{nsubj, obj, iobj, obl, cop, expl, xcomp} dependency are treated as an object (i.e., the word order is swapped).
\end{itemize}

That is, unless an argument is explicitly marked as a subject or marked topic, it is regarded as an object, which is compatible with the loose definition of object employed in the English experiment.

The second rule regarding the topicalization marker ``wa'' handles the topicalization phenomena.
Note that the Japanese language is topic-prominent~\cite{noda1996hatoga,teruya2004metafunctional,teruya2007systemic,fujihara-etal-2022-topicalization}, and a certain component of a sentence is frequently topicalized (i.e., moved to the initial part of the sentence with a special topicalization marker \texttt{TOP}).
For example, either the subject or object of a sentence (1) can be topicalized:

\resizebox{0.7\columnwidth}{!}{%
    \begin{dependency}[theme = default, arc angle = 30]
       \begin{deptext}[column sep=0.3em, row sep=0em, font=\large]
          \& PRON \& ADP \& NOUN \& ADP \& VERB \\
          (1) \& Watashi \& ga \& \textcolor{AccentBlue}{b\^oru} \& \textcolor{AccentBlue}{wo} \& \textcolor{AccentRed}{negata.} \\
          \& I \& \texttt{NOM} \& ball \& \texttt{ACC} \& throw. \\
       \end{deptext}
       \depedge[edge style={font=\Large}, label style={font=\normalsize}, edge height={0.5cm}]{2}{3}{case}
        \depedge[edge style={font=\Large}, label style={font=\normalsize}, edge height={1.5cm}]{6}{2}{nsubj}
       \depedge[edge style={font=\Large}, label style={font=\normalsize}, edge height={0.5cm}]{4}{5}{case}
       \depedge[edge style={font=\Large}, label style={font=\normalsize}, edge height={1cm}]{6}{4}{obj}
       \deproot[]{6}{root}
    \end{dependency} 
}

\noindent
The subject is topicalized in sentence (2), and the object is topicalized in sentence (3):

\resizebox{0.7\columnwidth}{!}{%
    \begin{dependency}[theme = default, arc angle = 30]
       \begin{deptext}[column sep=0.3em, row sep=0em, font=\large]
          \& PRON \& ADP \& NOUN \& ADP \& VERB \\
          (2) \& Watashi \& wa \& \textcolor{AccentBlue}{b\^oru} \& \textcolor{AccentBlue}{wo} \& \textcolor{AccentRed}{negata.} \\
          \& I \& \texttt{TOP} \& ball \& \texttt{ACC} \& throw. \\
       \end{deptext}
       \depedge[edge style={font=\Large}, label style={font=\normalsize}, edge height={0.5cm}]{2}{3}{case}
        \depedge[edge style={font=\Large}, label style={font=\normalsize}, edge height={1.5cm}]{6}{2}{nsubj}
       \depedge[edge style={font=\Large}, label style={font=\normalsize}, edge height={0.5cm}]{4}{5}{case}
       \depedge[edge style={font=\Large}, label style={font=\normalsize}, edge height={1cm}]{6}{4}{obj}
       \deproot[]{6}{root}
    \end{dependency} 
}

\resizebox{0.7\columnwidth}{!}{%
    \begin{dependency}[theme = default, arc angle = 30]
       \begin{deptext}[column sep=0.3em, row sep=0em, font=\large]
          \& PRON \& ADP \& NOUN \& ADP \& VERB \\
          (3) \& B\^oru \& wa \& watashi \& ga \& \textcolor{AccentRed}{negata.} \\
          \& Ball \& \texttt{TOP} \& I \& \texttt{NOM} \& throw. \\
       \end{deptext}
       \depedge[edge style={font=\Large}, label style={font=\normalsize}, edge height={0.5cm}]{2}{3}{case}
        \depedge[edge style={font=\Large}, label style={font=\normalsize}, edge height={1.5cm}]{6}{2}{nsubj:outer}
       \depedge[edge style={font=\Large}, label style={font=\normalsize}, edge height={0.5cm}]{4}{5}{case}
       \depedge[edge style={font=\Large}, label style={font=\normalsize}, edge height={1cm}]{6}{4}{nsubj}
       \deproot[]{6}{root}
    \end{dependency} 
} \\

\noindent
The topicalized component is typically ambiguous in terms of its grammatical case, and thus, the parser outputs were erroneous.
Such a \textit{marked} word order is beyond our interest since the Greenbergial correlations are generally on the canonical, \textit{unmarked} word order of language.
Thus, we did not modify the word order of such an explicitly topicalized word, even if it is seemingly an object of a verb.
For example, the topicalized object, ``B\^oru wa'' in Example (3), is no longer the target of \corr{<V, O>} swapping.

Another Japanese-specific concern is on a particular type of noun, called \textit{sa-hen} noun, which can behave as a verb with a special conjugation verb ``suru,'' e.g., ``y\^oyaku'' (\texttt{NOUN})$\rightarrow$``y\^oyaku-suru'' (\texttt{VERB}), like the English words``summary'' (\texttt{NOUN})$\rightarrow$``summar-ize'' (\texttt{VERB}).
However, the conjugation verb ``suru'' is sometimes omitted even when the sa-hen noun is used as a verb.
Such nouns are typically annotated as \texttt{NOUN} with objects in the Japanese UD:

\resizebox{0.9\columnwidth}{!}{%
\vspace{1cm}
\begin{dependency}[theme = default, arc angle = 30]
       \begin{deptext}[column sep=0.3em, row sep=0em, font=\large]
          PROPN \& ADP \& NOUN \& ADP \& NOUN \& ADP \& ADP \& NOUN \\
          Ken \& ga \& \textcolor{AccentBlue}{r\^esu} \& \textcolor{AccentBlue}{wo} \& \textcolor{AccentRed}{ketsuj\^o} \& to \& no \& uwasa... \\
          Ken \& \texttt{NOM} \& race \& \texttt{ACC} \& skip \& that \&  \& rumor... \\
       \end{deptext}
       \depedge[edge style={font=\Large}, label style={font=\normalsize}, edge height={0.5cm}]{1}{2}{case}
        \depedge[edge style={font=\Large}, label style={font=\normalsize}, edge height={1.5cm}]{5}{1}{nsubj}
       \depedge[edge style={font=\Large}, label style={font=\normalsize}, edge height={0.5cm}]{3}{4}{case}
        \depedge[edge style={font=\Large}, label style={font=\normalsize}, edge height={1cm}]{5}{3}{obj}
        \depedge[edge style={font=\Large}, label style={font=\normalsize}, edge height={0.5cm}]{5}{6}{case}
        \depedge[edge style={font=\Large}, label style={font=\normalsize}, edge height={1cm}]{5}{7}{case}
        \depedge[edge style={font=\Large}, label style={font=\normalsize}, edge height={1.5cm}]{8}{5}{nmod}
    \end{dependency} 
    }

\noindent
Here, ``\textcolor{AccentRed}{ketsuj\^o}'' (\textit{skip}) is annotated as a \texttt{NOUN} but can be regarded as a \texttt{VERB}, and the native Japanese validator indeed pointed out this should be included in the verb-object pairs.
Thus, we regarded \textit{sa-hen} nouns with either \deparc{nsubj, obj, iobj, obl, cop, expl, xcomp} dependent as verbs even when there is no conjugation verb.
With this rule, in the above example, ``\textcolor{AccentRed}{ketsuj\^o}'' is treated as a verb, and thus the position of its object ``\textcolor{AccentBlue}{r\^esu-wo}'' (\textit{race}-\texttt{ACC}) will be changed by the \corr{<V, O>} swapping algorithm.
 
\paragraph{Adpositions and Noun Phrases}
Japanese has a nominalizer, ``-no,'' which can convert any content word to a noun.
For example, a verb ``hataraku'' (\textit{work}) can be a noun with that nominalier ``hataraku-no'' (\textit{working}), but such nominalization is not reflected in the PoS tag of the nominalized words.
We regard the words nominalized by ``-no'' (tagged as \texttt{SCONJ}) as \texttt{NOUN} in this paper, and thus, the following sentence will also be a target of \corr{<Adp, NP>} swapping even though the head of the \deparc{nmod} dependency ``\textcolor{AccentRed}{karui}'' is \texttt{ADJ} rather than \texttt{NOUN}:

\resizebox{0.9\columnwidth}{!}{%
\begin{dependency}[theme = default, arc angle = 30]
       \begin{deptext}[column sep=0.3em, row sep=0em, font=\large]
          PRON \& ADP \& ADJ \& SCONJ \& ADP \& ADJ \& NOUN \& AUX \\
          \textcolor{AccentBlue}{Kare} \& \textcolor{AccentBlue}{no} \& \textcolor{AccentRed}{sugoi} \& \textcolor{AccentRed}{no} \& wa \& kashikoi \& tokoro \& da. \\
          He \& 's \& excellent \& -ness \& \texttt{TOP} \& smart \& aspect \& is.  \\
       \end{deptext}
       \depedge[edge style={font=\Large}, label style={font=\normalsize}, edge height={0.5cm}]{1}{2}{case}
        \depedge[edge style={font=\Large}, label style={font=\normalsize}, edge height={1.5cm}]{3}{1}{nmod}
       \depedge[edge style={font=\Large}, label style={font=\normalsize}, edge height={0.5cm}]{3}{4}{mark}
        \depedge[edge style={font=\Large}, label style={font=\normalsize}, edge height={1cm}]{3}{5}{case}
        \depedge[edge style={font=\Large}, label style={font=\normalsize}, edge height={1cm}]{7}{6}{acl}
        \depedge[edge style={font=\Large}, label style={font=\normalsize}, edge height={1.5cm}]{7}{3}{csubj}
        \depedge[edge style={font=\Large}, label style={font=\normalsize}, edge height={1cm}]{7}{8}{cop}
        \deproot[]{7}{root}
    \end{dependency} 
}

\paragraph{Copula and Predicate}
The \deparc{cop} dependency is attached only to an auxiliary verb ``desu'' in the original Japanese UD.
We increased the coverage of copula verb based on the following criteria:

\begin{itemize}
    \item \texttt{AUX} of ``dearu'' (\textit{is}), ``denai'' (\textit{is not}), ``dewanai'' (\textit{is not}),  ``janai'' (\textit{is not}), ``rash\^i'' (\textit{looks/seems/sounds like})  ``kamoshirenai'' (\textit{may be}) .
    \item \texttt{VERB} with ``iru'' (\textit{exist}), ``aru'' (\textit{exist}), or ``naru'' (\textit{become}) as its lexicon.
\end{itemize}

That is, in the following example, the original annotation on the left with the copula verb ``\textcolor{AccentRed}{dearu}'' is converted into the dependency graph on the right: \\

\resizebox{0.4\columnwidth}{!}{%
\begin{dependency}[theme = default, arc angle = 30]
       \begin{deptext}[column sep=0.3em, row sep=0em, font=\large]
          PRON \& ADP \& NOUN \& AUX \\
          \textcolor{AccentBlue}{Kore} \& \textcolor{AccentBlue}{wa} \& \textcolor{AccentRed}{jijitsu} \& \textcolor{AccentRed}{dearu.} \\
          This \& \texttt{TOP} \& fact \& is. \\
       \end{deptext}
       \depedge[edge style={font=\Large}, label style={font=\normalsize}, edge height={0.5cm}]{1}{2}{case}
        \depedge[edge style={font=\Large}, label style={font=\normalsize}, edge height={1cm}]{3}{1}{nsubj}
       \depedge[edge style={font=\Large}, label style={font=\normalsize}, edge height={0.5cm}]{3}{4}{aux}
        \deproot[]{3}{root}
    \end{dependency} 
}
\resizebox{0.4\columnwidth}{!}{%
    \begin{dependency}[theme = default, arc angle = 30]
       \begin{deptext}[column sep=0.3em, row sep=0em, font=\large]
          PRON \& ADP \& NOUN \& AUX \\
          \textcolor{AccentBlue}{Kore} \& \textcolor{AccentBlue}{wa} \& \textcolor{AccentRed}{jijitsu} \& \textcolor{AccentRed}{dearu.} \\
          This \& \texttt{TOP} \& fact \& is. \\
       \end{deptext}
       \depedge[edge style={font=\Large}, label style={font=\normalsize}, edge height={0.5cm}]{1}{2}{case}
        \depedge[edge style={font=\Large}, label style={font=\normalsize}, edge height={1cm}]{4}{1}{nsubj}
       \depedge[edge style={font=\Large}, label style={font=\normalsize}, edge height={0.5cm}]{4}{3}{cop*}
       \deproot[]{4}{root}
    \end{dependency} 
    }

Note that we only targeted the cases where the copula verb has a \deparc{nsubj} dependent since the corresponding construction in English, i.e.,  a sentence ``A is B.'' with the omission of ``A,'' is very rare.

\paragraph{Auxiliary and Verb}
While auxiliary words are swapped with an entire verb phrase rather than a single verb in the English implementation of \corr{<Aux, V>} swapping, the Japanese implementation only swaps a single verb.
This is because Japanese auxiliary verbs are typically analyzed as affixes, and thus separating them from the verb modifies the language beyond simply breaking the Greenbergian correlation.
Taking the sentence in Table \ref{tab:japanese_examples} as an example, the auxiliary ``\textcolor{AccentBlue}{teiru}'' is moved immediately before the verb ``\textcolor{AccentRed}{tsudui},'' rather than the initial position of the sentence, regarding the whole descendants of the verb  (``Ichigo no kisetsu ga shichigatsu kara hachigatsu made'') in the \corr{<Aux, V>} variant.

\paragraph{Noun and Genitive}
We identified the genitive constructions as follows:
\begin{itemize}
    \item A \deparc{nmod} dependency to a noun phrase.
    \item The dependent has either particle of ``no,'' ``ga,'' or ``tsu.'' 
\end{itemize}
We exclude some exceptional constructions; for example, we did not swap the expression ``X-no y\^o na'' to be ``y\^o X-no na.''
We also considered the nominalization in identifying a noun, as explained in the \corr{<Adp, NP>} swapping.

\section{General Swapping Algorithm}

Algorithm 1 below is the basic form of the depth-first swapping algorithm. This basic algorithm was modified to handle the specific of each language and correlation pair as described in \cref{app:details}.

\begin{algorithm}
\footnotesize
\caption{Swapping Greenbergian correlation pairs in a sentence}
\begin{algorithmic}[1]  %
\Func{Swap}{sentence $s$, UD parse $p$, Correlation pair \corr{<X, Y>}}  %
    \State $stack \gets [root]$
    \State $visited \gets set()$

    \While{$stack$ is not empty}
        \State $node \gets \Call{Pop}{stack}$
        \If{$node$ is not in $visited$}
            \State \Call{AddToVisited}{$visited$, $node$}
            \For{each child $c$ of $node$ in the parse $p$ of $s$}
                \If{$node$ is verb-patterner $X$ and $c$ is object-patterner $Y$}
                    \State \Call{SwapPair}{$node$, $c$, $s$, $p$}
                \EndIf
                \If{$c$ is not in $visited$}
                    \State \Call{Push}{$stack$, $c$}
                \EndIf
            \EndFor
        \EndIf
    \EndWhile
    \State \Return $s$
\EndFunc
\end{algorithmic}
\label{alg:swap}
\end{algorithm}

\section{Additional Annotation Guidelines}\label{app:annotation}

The 5-point Likert scale used to evaluate the validity of swapped sentences is given below:

\begin{enumerate}
\setlength{\parskip}{0cm}
\setlength{\itemsep}{0.1cm}
    \item All or most swaps have serious errors
    \item A few serious errors or several small errors
    \item A few small errors
    \item A minor error or less likely but valid changes
    \item Perfect
\end{enumerate}

\section{Details on Experimental settings}\label{app:lm}
\paragraph{Language Models}
All models are trained using the HuggingFace library \citep{wolf-etal-2020-transformers}.
For GPT-2 small model, sub-word tokenization is implemented by Byte-Pair Encoding (BPE) algorithm \citep{sennrich-etal-2016-neural} with a vocabulary size of 32,000. 
For LTG-BERT, we adopted the same WordPiece tokenizer with a vocabulary size of $2^{14} = 16384$ as in the original implementation \citep{samuel2023trained}, only removing special characters \textit{<TAB>} and \textit{<PAR>} as it doesn't apply to Wiki-40B text format.

\paragraph{Stanza Parsers}
We use Stanza~\cite{qi2020stanza} version 1.5.1 and 1.6.1 based on the UD 2.0 formalism~\cite{nivre-etal-2020-universal} for English and Japanese, respectively. 
For Japanese, we used a long-unit-word (LUW) parser (\url{https://github.com/UniversalDependencies/UD_Japanese-GSDLUW}) which is more compatible with the syntactic UD scheme \citep{omura-etal-2021-word} rather than the default, short-unit-word (SUW) parser which is better for morphological analysis \citep{tanaka-etal-2016-universal,murawaki2019definition,pringle2022thoughts}.

\end{document}

%% file: macros.tex
\newcommand{\mymacro}[1]{{\color{MacroColor} #1}}

\newcommand{\newterm}[1]{\textbf{#1}}

\def\cites#1{\citeauthor{#1}'s (\citeyear{#1})}

\newcommand{\paroutline}[3][false]{%
  \ifnum\pdfstrcmp{#1}{true}=0
    #3
  \else
    [\textit{\textcolor{DiverseMagenta}{#2}}] \textcolor{AccentBlue}{#3}
  \fi
}

\newcommand{\postag}[1]{\textsf{#1}}
\newcommand{\deparc}[1]{\texttt{#1}}
\newcommand{\corr}[1]{\textit{#1}}
\newcommand{\vlm}[1]{\textsc{#1}}

\newcommand{\negterm}[1]{{\mymacro{ {\raise.17ex\hbox{$\scriptstyle\sim$}} #1}}}

\newcommand{\ignore}[1]{}
\newcommand{\expandLater}[1]{}

\def\1{\mathbf{1}}